\documentclass[10pt,twocolumn,letterpaper]{article}

\usepackage[pagenumbers]{cvpr} 

\usepackage{booktabs, makecell}
\usepackage{cellspace}
\usepackage{epsfig}
\usepackage{graphicx}
\usepackage{amsmath}
\usepackage{amssymb}
\usepackage{multirow}
\usepackage{wrapfig}
\usepackage{tikz}
\usepackage{colortbl}
\usepackage{algpseudocode}
\usepackage[innercaption]{sidecap}
\usepackage{ifthen}
\usepackage{nicefrac}
\usepackage{dsfont}
\usepackage{algorithm}
\usepackage{float}
\usepackage{makecell}
\usepackage{lipsum}
\usepackage{bm}

\usepackage{pgfplots}
\pgfplotsset{compat=1.11,
    /pgfplots/ybar legend/.style={
    /pgfplots/legend image code/.code={
       \draw[##1,/tikz/.cd,yshift=-0.25em]
        (0cm,0cm) rectangle (3pt,0.8em);},
   },
}

\definecolor{Gray}{gray}{0.9}
\def\red#1{\textcolor[rgb]{1,0,0}{#1}}

\definecolor{caribbeangreen}{rgb}{0.0, 0.8, 0.6}
\definecolor{mygreen}{rgb}{0.0, 0.8, 0.0}
\definecolor{deepGreen}{RGB}{0,153,0}
\definecolor{orange}{RGB}{250,153,0}

\usepackage{soul}
\DeclareRobustCommand{\hlgood}[1]{{\sethlcolor{SpringGreen}\hl{#1}}}
\DeclareRobustCommand{\hlbad}[1]{{\sethlcolor{Melon}\hl{#1}}}

\newcommand{\keypoint}[1]{\vspace{0.08cm}\noindent\textbf{#1}\quad}
\newcommand{\cut}[1]{}
\newcolumntype{g}{>{\columncolor{Gray}}c}

\algnewcommand\algorithmicforeach{\textbf{for each}}
\algdef{S}[FOR]{ForEach}[1]{\algorithmicforeach\ #1\ \algorithmicdo}

\newcommand{\scriptveryshortarrow}[1][3pt]{{%
    \hbox{\rule[\scriptratio\dimexpr\fontdimen22\textfont2-.2pt\relax]
               {\scriptratio\dimexpr#1\relax}{\scriptratio\dimexpr.4pt\relax}}%
   \mkern-4mu\hbox{\let\f@size\sf@size\usefont{U}{lasy}{m}{n}\symbol{41}}}}

\definecolor{cvprblue}{rgb}{0.21,0.49,0.74}
\usepackage[pagebackref,breaklinks,colorlinks,allcolors=cvprblue]{hyperref}

\def\paperID{07651} 
\def\confName{CVPR}
\def\confYear{2025}

\usepackage{orcidlink}

\usepackage[capitalize]{cleveref}
\crefname{section}{Sec.}{Secs.}
\Crefname{section}{Section}{Sections}
\Crefname{table}{Table}{Tables}
\crefname{table}{Tab.}{Tabs.}

\usepackage[symbol]{footmisc}

\title{\vspace{-2mm}Sketch Down the FLOPs: Towards Efficient Networks for Human Sketch\vspace{-2mm}}

\begin{document}

\author{
Aneeshan Sain\textsuperscript{1}  \hspace{.2cm} 
Subhajit Maity\textsuperscript{2}\thanks{Work done as an intern at SketchX before joining UCF.} \hspace{.3cm}
Pinaki Nath Chowdhury\textsuperscript{1} \hspace{.2cm}  
Subhadeep Koley\textsuperscript{1} \hspace{.2cm}   \\
Ayan Kumar Bhunia\textsuperscript{1} \hspace{.3cm}   
Yi-Zhe Song\textsuperscript{1}  \\
\textsuperscript{1} SketchX, CVSSP, University of Surrey, United Kingdom.\\
\textsuperscript{2} Department of Computer Science, University of Central Florida, United States.\\
{\tt\small \{a.sain, p.chowdhury, s.koley, a.bhunia, y.song\}@surrey.ac.uk; Subhajit@ucf.edu} 
}
\maketitle

\begin{abstract}
As sketch research has collectively matured over time, its adaptation for at-mass commercialisation emerges on the immediate horizon. Despite an already mature research endeavour for photos, there is no research on the efficient inference specifically designed for sketch data. In this paper, we first demonstrate existing state-of-the-art efficient light-weight models designed for photos do not work on sketches. We then propose two sketch-specific components which work in a plug-n-play manner on any photo efficient network to adapt them to work on sketch data. We specifically chose fine-grained sketch-based image retrieval (FG-SBIR) as a demonstrator as the most recognised sketch problem with immediate commercial value. Technically speaking, we first propose a cross-modal knowledge distillation network to transfer existing photo efficient networks to be compatible with sketch, which brings down number of FLOPs and model parameters by $97.96$\% percent and $84.89$\% respectively. We then exploit the abstract trait of sketch to introduce a RL-based canvas selector that dynamically adjusts to the abstraction level which further cuts down number of FLOPs by two thirds. The end result is an overall reduction of $99.37$\% of FLOPs (from $40.18$G to $0.254$G) when compared with a full network, while retaining the accuracy ($33.03$\% vs $32.77$\%) -- finally making an efficient network for the sparse sketch data that exhibit even fewer FLOPs than the best photo counterpart.
\end{abstract}


\vspace{-4mm}
\section{Introduction}
\vspace{-0.2cm}
\label{sec:intro}

A significant movement in the late computer vision literature has been that of model efficiency. This effort has been largely driven by the on-the-edge deployments of vision models. With great strides made on networks specifically tackling photo data (MobileNet \cite{sandler2018mobilenetv2}, EfficientNet \cite{tan2019efficientnet}, etc), there however has been surprisingly no work targeting human sketches~\cite{xu2022deep}. This is despite the abundance of work specifically addressing sketch data~\cite{chowdhury2023what} and its unique traits: sequential~\cite{ha2018neural}, abstract~\cite{koley2023picture}, or stroke-wise~\cite{bhunia2023sketch2saliency}, style-diversity~\cite{sain2021stylemeup}, and  data-scarcity~\cite{bhunia2020sketch, pang2019generalising, dutta2019semantically}. This is particularly disappointing for the problem of fine-grained sketch-based image retrieval (FG-SBIR), which is perhaps the single-most studied sketch task \cite{yu2016sketch,umar2019goal,pang2019generalising,sain2023clip,sain2023exploiting,bhunia2021more,bhunia2022sketching,PartialSBIR} that has already matured for at-mass commercial adoption, and is our problem of choice in this paper.

\begin{figure}[!tb]
\centering
\includegraphics[width=\linewidth]{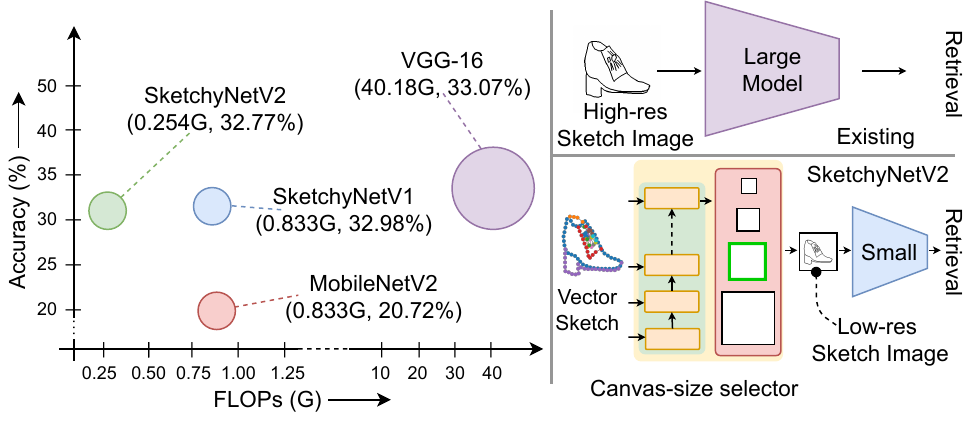}
\vspace{-4mm}
\caption{Our SketchyNetV1 compresses \textit{existing} heavy FG-SBIR networks to deliver smaller models. Further enhanced via a canvas-selector module our SketchyNetV2 model minimises sketch-resolution dynamically to reduce FLOPs.}
\label{fig:teaser}
\vspace{-6mm}
\end{figure}

One would think that state-of-the-art models designed for photos would work as-is for sketches, treating sketch data as raster images (which sketch is not). As our first contribution, we preempt this thought via a pilot study to prove its false nature. In particular, we borrow MobileNetv2 \cite{sandler2018mobilenetv2} as a backbone network to construct a typical triplet loss based network for FG-SBIR. Results show a reduction of almost 37\% (\textit{relative}) in retrieval accuracy when compared with a typical VGG-16 backbone \cite{SimonyanZ14a}. The focus of this paper is however not to design entirely new sketch-specific efficient networks from scratch -- lots of good efficient photo networks already exist. Our goal instead is to adapt existing photo networks to work with sketch, by introducing generic sketch-specific designs that work in a plug-n-play manner. 

We first tackle the problem highlighted in the said pilot study -- to make existing photo efficient networks perform on the same level as large networks on FG-SBIR. We argue the reason behind why the likes of MobileNetV2 did not work off-the-shelf is that the limited set of parameters could not accommodate the fine-grained nature of sketches (i.e., sketches are highly expressive in capturing fine-grained visual details \cite{xu2022deep}) for sketch and photo matching. For that, we introduce a \textit{knowledge distillation} (KD) paradigm~\cite{romero2015fitnets} to adapt this sketch trait onto efficient photo networks. The core idea is to induce strong {semantic} knowledge between a sketch and its photo, in the embedding space of a student, from a deep and high performing teacher. This particularly helps the efficient model to achieve on par performance with the deep variant without having a large parameter size. However, ours being a cross-modal retrieval, naively adapting standard KD paradigms involving logit distillation \cite{ba2014deep} would be sub-optimal. We therefore aim to preserve cross-modal pairwise distances among instances, and in turn the structural consistency of the teacher's discriminative embedding space in the student. The result is already an efficient FG-SBIR model (which we term SketchyNetV1) that reduces the number of FLOPs from 40.18G (VGG-16 based model\cite{sain2021stylemeup}) to 0.833G (our MobileNetV2-based) -- a roughly 48 times reduction (\cref{fig:teaser} left).

This abstract nature of human sketch~\cite{chowdhury2023democratising} can also be exploited, to further reduce the FLOPs. The hypothesis is that since sketches are essentially abstract depictions of images, they should carry a similar level of semantic information even when rendered at much lower resolutions, and that more abstract a sketch is the more information compact it is (i.e., can sustain even lower resolutions). For that, we conducted another pilot study (\cref{sec:Pilot}), where we show unlike photos, sketches can sustain retrieval accuracy with much lower resolutions (4\% vs 15\% at 32$\times$32), and ~30\% of those query sketches achieving perfect retrieval at full-resolution, when downsized to 32$\times$32 can still retrieve accurately (\cref{fig:pilot_plot} right). This motivated us to design a canvas selector that can choose the right resolution \textit{on-the-fly} to render a sketch before feeding to the network. This is thanks to another unique trait of sketch data -- they come in vector format and can be freely rendered to any resolution without extra overhead. Accomplishing this however is non-trivial since this selection process of canvas-size involves a discrete decision, thus invoking non-differentiability.\cut{ in the network.} We thus leverage reinforcement learning~\cite{bhunia2022sketching} to bypass these issues and train an abstraction-aware \textit{canvas-size selector} as a policy network for our model, which estimates the optimal canvas-size required for the input sketch. The FG-SBIR model, acting as a critic, calculates reward for the canvas-selector, balancing between accuracy and FLOPs, that trains the policy network. With this further trait of sketches addressed, our final model (SketchyNetV2) was able to further reduce FLOPs by a factor of 3.28 (from 0.833G to 0.254G), while retaining a retrieval accuracy similar to the full model (33.03 vs 32.77), exactly as one would imagine sketch efficient networks to entail (sketches being sparse). 

In summary, our contributions are: (a) the first investigation towards efficient sketch networks from existing photo ones (b) a knowledge distillation framework specifically designed to cater to the fine-grained nature of sketches, (c) a canvas-size selector that caters to the abstract nature of sketches, which further reduces FLOPs by two-thirds. Extensive experiments show our method to perform at par with existing state-of-the-arts, while being superiorly efficient. 

\vspace{-0.15cm}
\section{Related Work}
\label{sec:related}
\vspace{-0.20cm}
                      
\keypoint{Fine-grained SBIR:} 
Compared to category-level SBIR \cite{liu2017deep, collomosse2019livesketch, dey2019doodle, sketch3T}, Fine-Grained SBIR (FG-SBIR) focuses on retrieving a single photo instance from a category-specific gallery based on a query sketch. Introduced as a deep triplet-ranking Siamese network \cite{yu2016sketch}, FG-SBIR has evolved with attention-based modules \cite{song2017deep}, textual tags \cite{song2017fine}, hybrid cross-domain generation \cite{pang2017cross}, hierarchical co-attention \cite{sain2020cross}, and various pre-training strategies like mixed modal jigsaw solving ~\cite{pang2020solving}, and reinforcement learning~\cite{bhunia2020sketch}. Addressing sketch-specific traits like style diversity \cite{sain2021stylemeup}, data scarcity \cite{bhunia2021more}, and sketch stroke redundancy \cite{bhunia2022sketching}, improvements have been made for enhanced retrieval \cite{bhunia2022sketching}. Recently it was extended to scene-level (retrieve a scene-image, given a scene-sketch), employing cross-modal region associativity~\cite{PartialSBIR} and enhanced using text-queries~\cite{chowdhury2023scenetrilogy}. However, sketch-specific computational efficiency has largely been ignored -- which we address here.

\noindent \textbf{Towards Computational Efficiency:} 
Addressing large-scale deployment, research emerging in computational efficiency~\cite{alwani2022decore}, include network pruning~\cite{luo2017thinet}, quantisation~\cite{gupta2015deep}, binarisation~\cite{courbariaux2015binaryconnect}, knowledge distillation~\cite{hinton2015distilling}, input-size optimisation~\cite{talebi2021learning}, or their combinations~\cite{alwani2022decore}. Network \textit{pruning}~\cite{han2015deep,wang2016cnnpack} involves discovering and dropping low impact weights or channels of a pre-trained large network, while retaining its original accuracy. Contrarily, \textit{ quantisation} directly reduces the bit-width of parameter-values~\cite{gupta2015deep} and gradients~\cite{zhou2016dorefa}, by replacing floating-point computations with faster and cheaper low-precision fixed-point numbers~\cite{lin2016fixed}. Its extreme version, \textit{binarisation}~\cite{courbariaux2015binaryconnect}, binarises weights and activations. \textit{Knowledge Distillation} aims at transferring knowledge of a large pre-trained \textit{teacher} network to a smaller \textit{student} for cost-effective deployment. Existing techniques leverage output logits~\cite{ba2014deep}, hidden layers~\cite{romero2015fitnets},  attention-maps~\cite{zagoruyko2017KDattention}, or {neuron selectivity} pattern~\cite{huang2017like} of pre-trained teachers for the same. Self-distillation~\cite{bagherinezhad2018label} employs the same network for both student and teacher models, to further reduce compute. \textit{Input-resolution optimisation} on the other hand involves replacing a high-resolution input image with its down-scaled low-resolution version~\cite{talebi2021learning}, to reduce overall compute \cite{xu2022smartadapt}, while retaining  accuracy. Relevant works explored dynamic usage of high-resolution data~\cite{uzkent2020learning} or  patch proposals for strategic image-cropping~\cite{wang2020glance}, where the learnable resizer \cite{talebi2021learning} operates on the original image resolution. All such methods however cater to compressing image-specific models, which if used off-the-shelf for FG-SBIR would fail to address the sparse manner in which sketches are represented \cite{bhunia2021vectorization}. We thus propose a \textit{sketch-specific} light-weight FG-SBIR model, tackling sparsity in sketches.

\noindent \textbf{Reinforcement Learning in Vision (RL):} 
Applying RL~\cite{kaelbling1996reinforcement} to computer vision problems~\cite{liang2017deep, wang2019reinforced} is commonplace of late. Whenever quantifying the \textit{goodness} of the network's state is non-differentiable, like selecting which regions to crop in an image~\cite{li2018a2}, RL comes in handy. In sketch community, RL has found use in retrieval~\cite{bhunia2020sketch, bhunia2021more}, modelling sketch abstraction~\cite{muhammad2018learning, umar2019goal}, designing competitive sketching agents~\cite{bhunia2020pixelor} and as a stroke-selector~\cite{bhunia2022sketching}. Here, we leverage RL for optimising a canvas-size selector for a computationally efficient retrieval model.

\begin{table}[t]
\vspace{-0.2cm}
\scriptsize
\setlength{\tabcolsep}{6pt}
\renewcommand{\arraystretch}{01.0}
    \centering
    \caption{Performance \& computational cost for various networks}
    \vspace{-0.35cm}
    \begin{tabular}{lcccc}
         \toprule
         Model & FLOPS & Parameters & Time (ms) & Acc@1 \\
         \midrule 
         VGG-19~\cite{SimonyanZ14a} &  51.06G & 20.02M & 101 & 33.63\\
         VGG-16~\cite{SimonyanZ14a} &  40.18G & 14.71M & 084 & 33.03\\
         ResNet-101 \cite{HeResNet} &  20.50G & 42.50M & 155 & 21.77\\
         DenseNet-201~\cite{huang2017densely} &  11.40G & 18.09M & 335 &  23.62\\
         ResNet-50 \cite{HeResNet} &  10.76G & 23.51M & 086 & 21.32\\
         InceptionV3~\cite{yu2016sketch} &  07.72G & 21.79M & 134 & 24.27\\
         \midrule
         ResNet-18~\cite{HeResNet} & 4.76G & 11.18M & 036 & 20.72\\
         EfficientNet~\cite{tan2019efficientnet} &  1.05G & 04.01M & 112 & 22.47\\
         MobileNetV2~\cite{sandler2018mobilenetv2} &  0.83G & 02.22M & 070 & 20.85\\
         \bottomrule
    \end{tabular}
    \vspace{-0.6cm}
    \label{tab:pilot1}
\end{table}

\vspace{-2mm}
\section{Pilot Study}
\label{sec:Pilot}
\vspace{-3mm}
\keypoint{Backbone Network for FG-SBIR:}
Amongst popular backbones used in recent FG-SBIR literature ~\cite{bhunia2020sketch, bhunia2021more, sain2020cross, sain2021stylemeup}, are VGG-16~\cite{simonyan2015very} and Inception-V3~\cite{yu2016sketch}. Others like ResNet18~\cite{HeResNet} and EfficientNet~\cite{tan2019efficientnet}, though cheaper and prevalent in vision~\cite{HeResNet} are rarely used in FG-SBIR. Exploring their potential in FG-SBIR, we train a \textit{baseline} FG-SBIR model, plugging in popular ones as backbones for our study. \Cref{tab:pilot1} reports their parameter-count, FLOPs, time for feature-extraction per sample and Top-1 accuracy (\cref{sec:expt}) on standard QMUL-ShoeV2 dataset~\cite{yu2016sketch}. Despite scoring more than VGG-16 on image-classification \cite{vatathanavaro2018white}, ResNet50 \cite{HeResNet} performs poorly -- indicating that modelling fine-grained sketch-photo association needs networks with higher FLOPs. Despite delivering better accuracies, heavier networks cost far more resources. We thus aim for a \textit{lighter} model that achieves \textit{accuracy like a larger} one.

\begin{figure}[t]
\begin{center}
\includegraphics[width=0.9\linewidth]{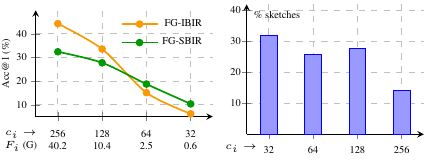}
\end{center}
\vspace{-0.7cm}
\caption{{Pilot study for varying canvas-size (see text above).}}
\label{fig:pilot_plot}
\vspace{-0.6cm}
\end{figure}

\keypoint{Dynamic Sketch-canvas-size:}
Unlike photos that hold pixel-dense information, sketches are sparse black and white lines. This begs the question, if a sketch rendered at a higher resolution with added computational burden would convey any extra semantic information than at a lower one. To answer this, we focus on vector-format of sketches \cite{bhunia2021more}, where every sketch vector $s_v$ can be characterised as a sequence of points ($v_1$,$v_2$, $\cdots$ ,$v_\text{T}$), and rendered to any canvas-size ($c$) as a raster-image ($s_r$) via \emph{rasterisation} $\mathcal{R}^c (\cdot): s_v \rightarrow s_r^c$ where $s_r^c \in \mathbb{R}^{c \times c \times 3}$. A \textit{single baseline} FG-SBIR model (VGG-16) is trained and tested on full-resolution images, and sketches rendered at varying canvas-sizes from  $C$=$\{32\times 32, \cdots, 256\times 256\}$. Comparing sparse sketches \emph{versus} photos, we train an equivalent fine-grained image-based image retrieval (FG-IBIR) where only the query against a positive photo is constructed using random augmentation \cite{chen2020simple} and resized to corresponding canvas-size $c_i$. Gallery photo-features for both are kept pre-computed. While \textit{FG-IBIR} accuracy falls rapidly (\cref{fig:pilot_plot} left), FG-SBIR stays relatively stable against decreasing canvas-sizes, as photos (unlike sketches) containing pixel-dense perfect information, lose a lot of it while down-scaling. Furthermore, \textit{positive} accuracy of FG-SBIR at $32\times32$, shows some sketches to hold sufficient semantic information for retrieval even at minimal canvas-size. \cref{fig:pilot_plot} (right) shows overall QMUL-ShoeV2\cite{yu2016sketch} statistics where each bar represents the \textit{smallest} canvas-size, at which \textit{bar-height} percentage of sketches achieve perfect retrieval (taking total sketches achieving perfect retrieval at 256$\times$256 as 100\%). This infers that every sketch has its own optimal canvas, and fixing it (\cref{fig:pilot_plot} left), for the entire dataset would always be sub-optimal. Moreover, the optimal canvas-size for a sketch, depends on its extent of sparsity, for which neither hard-ground-truth exists, nor is it knowable from the user's end. Finally, brute force discovery at every size being infeasible during inference, motivates us to design a \textit{learnable} adaptive canvas-size selector for sketches.

\vspace{-4mm}

\begin{figure*}[t]
\centering
\includegraphics[width=\linewidth]{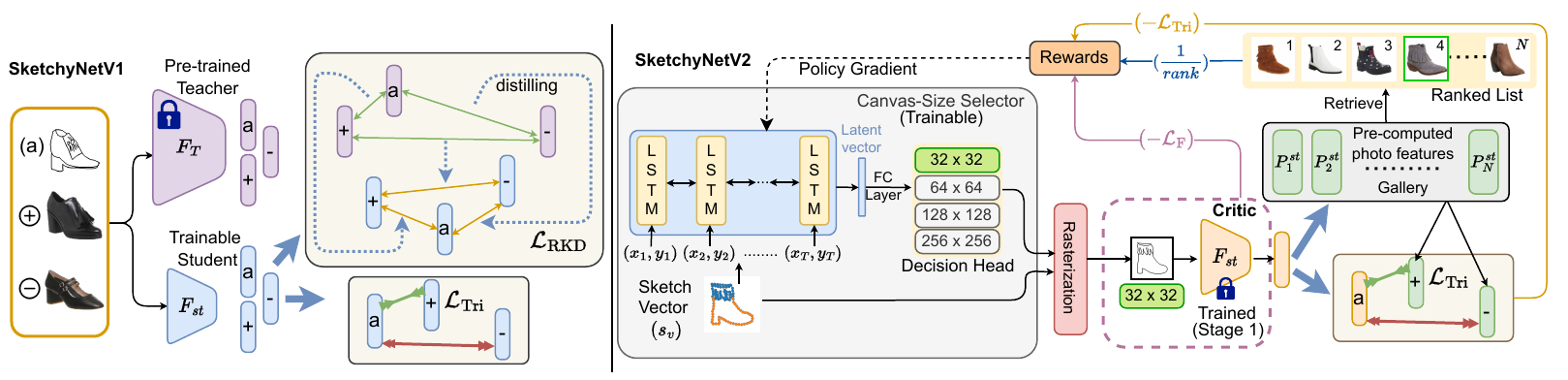}
\vspace{-7mm}
\caption{
Two stage framework. (Left) SketchyNetV1: A smaller \textbf{st}udent network ($F_{st}$) is trained from a larger pre-trained \textbf{t}eacher ($F_T$) via KD (\S \ref{sec:student_model}). (Right) We progress to SketchyNetV2 by training a canvas-size selector directed by objectives of increasing performance and reducing compute. It takes sketch as a vector ($s_v$) and aims to predict an optimal canvas-size at which the sketch is rasterised ($s_r$). $s_r$ when processed by $F_{st}$ minimises overall FLOPS, while retaining accuracy of corresponding full-resolution sketch-image. 
}
\label{fig:network}
\vspace{-4.0mm}
\end{figure*}

\section{Methodology}
\label{sec:methodology}
\vspace{-1.5mm}
\noindent \textbf{Overview:} Our pilot study motivates us to extend a standard FG-SBIR framework towards computational efficiency by first introducing \emph{SketchyNetV1} via knowledge distillation, and then taking it further to \emph{SketchyNetV2} via dynamic sketch canvas-size selection. Importantly, these are done without hurting the original performance of large-scale slower baseline model. In particular, there are two phases of improvement for computational efficiency. Firstly, we train a \textbf{t}eacher model $F_{T}$ as a standard baseline FG-SBIR model (\cref{sec:Pilot}). Once trained, its rich semantic knowledge is transferred to a smaller \textbf{st}udent network $F_{st}$ via a knowledge distillation (KD) paradigm, which we name as \emph{SketchyNetV1}. Next, we train a canvas-size selector that adaptively decides optimal canvas-size or resolution per sketch for \emph{SketchyNetV2} that would be sufficient to preserve the retrieval performance while reducing the number of FLOP operations in feature extraction significantly.

\noindent \textbf{\textit{Baseline} FG-SBIR model: } Keeping aside complex pre-training \cite{pang2020solving} or joint-training \cite{bhunia2021more}, we focus on a simple triple-branch state-of-the-art Siamese network~\cite{yu2016sketch} which remains a strong baseline so far \cite{sain2020cross,sain2021stylemeup,bhunia2021more}. A VGG-16 \cite{simonyan2015very} pre-trained on ImageNet~\cite{russakovsky2015imagenet} acts as the shared backbone for each branch. Given an input image $I$, a $d$ dimensional feature embedding is extracted from the backbone network $\mathcal{F}$ using global average pooling followed by  $l_2$ normalisation as $f_I \in \mathbb{R}^d$. The model is trained on triplets comprising features of an anchor sketch ($f_s$), its matching positive ($f_{p}$) photo and a random non-matching negative photo ($f_{n}$), via a triplet loss \cite{weinberger2009distance}. Triplet loss ($\mathcal{L}_{\text{Tri}}$) aims to minimise the distance between sketch and positive photo, while maximising the same from its negative (n). With $\delta(a,b)$ = $||a-b||^2$, as a distance function and a margin $m > 0$, we have:

\vspace{-3mm}
{\small
\begin{equation}
    \begin{aligned}
    \mathcal{L}_{\text{Tri}} = \max \{0, m + \delta(f_s, f_{p}) - \delta(f_{s}, f_{n})\}
    \end{aligned}
    \label{equ:baseTrip}
\end{equation}
}

\vspace{-1mm}
\subsection{SketchyNetV1: Smaller Model via KD}
\label{sec:student_model}
\vspace{-1mm}
Based on our pilot study (\cref{sec:Pilot}), if we naively train from an ImageNet pre-trained standard smaller network (e.g. MobileNetV2 \cite{sandler2018mobilenetv2} -- usually popular for edge devices), with simple supervised triplet loss (Eq.\ref{equ:baseTrip}) it can hardly reach the accuracy (\Cref{tab:pilot1}) obtained by a large network like VGG-16. Therefore apart from training via a supervised loss on the sketch-positive-negative triplet ($\mathcal{L}^{st}_\text{Tri}$ as in Eq.\ref{equ:baseTrip}), we aim to use the pre-trained larger teacher (e.g. VGG-16) for additional supervision. Accordingly, we adhere to Knowledge Distillation (KD) paradigm \cite{romero2015fitnets}, a conventional strategy for model compression, to deliver a light-weight student model from the larger teacher network \cite{romero2015fitnets}.
Applying KD to our FG-SBIR paradigm is however non-trivial. Unlike traditional KD methods, which often involve logit distillation for classification tasks, our scenario involves cross-modal retrieval. Here, the output is a continuous $d$-dimensional feature within a joint-embedding space. Furthermore, straightforward regression between teacher and student features for sketch and photo branches may encounter compatibility issues, given the disparity in the embedding spaces. Addressing this, if the dimensions of teacher and student embeddings differ, an additional feature transformation layer is necessary for alignment \cite{romero2015fitnets}. Overcoming these issues, we aim to distil the inter-feature distances from respective teacher networks to the student thus preserving structural orientation of features in the teacher's embedding space to that in the student, during KD (\cref{fig:network} (left)). 

Given features of sketch-positive-negative triplet of teacher ($f_s^T, f_p^T, f_n^T$), we calculate the inter-feature Euclidean distances in the output representation space of teacher ($F_T$) using $\delta(\cdot,\cdot)$ (Eq.\ref{equ:baseTrip}), as 
$d_{sp}^T$ = $\delta(f_s^T, f_p^T)$ ; $d_{sn}^T$ = $\delta(f_s^T, f_n^T)$ and $d_{pn}^T$ = $\delta(f_{p}^T, f_{n}^T)$.
Passing the same samples through the student, we similarly obtain $d_{sp}^{st}, d_{sn}^{st}$ and $d_{pn}^{st}$ using ($f_s^{st}, f_p^{st}, f_n^{st}$).
A smooth $l_1$ loss distils the $l_2$ distances from the teacher to the student. With $\mathcal{L}_\delta$ as a Huber-loss~\cite{huang2021robust}, relational distillation loss for sketch-photo pair between teacher and student is given as, 

\vspace{-2mm}
{\small
\begin{equation}
\label{equ:loss_rkd}
\begin{aligned}
    \mathcal{L}_\text{RKD}^{sp} &= 
    \mathcal{L}_\delta(d_{sp}^T,d_{sp}^{st})    \quad \text{where}, \\
    \mathcal{L}_\delta(a,b) &=  \begin{cases}
    \frac{1}{2} (a - b)^2  & \text{if $|a - b| < \beta$} \\
    \beta(|a - b| - \frac{1}{2}\beta) & \text{otherwise}
    \end{cases}
\end{aligned}
\end{equation}
}

Similarly, we obtain $\mathcal{L}_\text{RKD}^{sn}$ and $\mathcal{L}_\text{RKD}^{pn}$, to compute distillation loss, \small{$\mathcal{L}_\text{RKD} = \mathcal{L}_\text{RKD}^{sp} + \mathcal{L}_\text{RKD}^{sn} + \mathcal{L}_\text{RKD}^{pn}$}. With hyperparameter $\lambda$, overall student-training objective becomes:

\vspace{-2mm}
{\small
\begin{equation}
\label{equ:loss_trn}
\begin{aligned}
    \mathcal{L}_\text{trn}^{st} = \lambda\mathcal{L}_\text{Tri} + (1-\lambda) \mathcal{L}_\text{RKD}
\end{aligned}
\vspace{-1mm}
\end{equation}
}

\noindent The student trains on varying sketch-canvas-sizes ($C$), using their respective full resolution for teacher, to ensure scale-invariance as \small{$\mathcal{L}_\text{trn}^{st*} = \frac{1}{4}\sum_{i=1}^4 \mathcal{L}_\text{trn}^{st(c_i)}$}. However, as optimal canvas-size varies across sketches (\cref{fig:pilot_plot}), we need a learnable canvas-size selector, dynamically predicting the optimal size for each sketch. 

\vspace{-1mm}
\subsection{SketchyNetV2: Adaptive canvas-size Selector}
\label{sec:rl_policygrad}
\vspace{-1mm}
\noindent \textbf{Overview:} 
Understanding that an input at a lower resolution, would incur lesser FLOPs on evaluation than a higher one, we upgrade \textit{SketchyNetV1} ($F_{st}$) to a more computationally efficient \textit{SketchyNetV2} by introducing an adaptive canvas-size selector. It is designed in the vector modality as it is much cheaper \cite{xu2021multigraph} than its raster-image counterpart, and can dynamically encode the varying abstraction level of sketches. Furthermore, predicting the optimal canvas-size from vector modality, would make the corresponding rasterisation operation $\mathcal{R}^c(\cdot)$ at lower sizes cheaper, besides reducing FLOPs in feature extraction. Taking sketch vector $s_v$ as input, $\psi_C$ predicts the probability $p(c|s_v)$ of choosing one of $K$ discreet canvas-sizes from a set of $C$ = $\{32\times32, \cdots, 256\times256\}$. $s_v$ is then rasterised to $s_r^c \in \mathbb{R}^{c \times c \times 3}$ with the optimal canvas-size $c$, as $\mathcal{R}^c (\cdot): s_v \rightarrow s_r^c$, and fed to $F_{st}$ for retrieval.

Selecting an optimal canvas-size however is an ill-posed problem. Firstly, there are no explicit labels representing the optimal canvas-size for a sketch. Secondly, annotating the optimal canvas-size for the whole dataset via brute-force iteration is computationally impractical. We therefore use the pre-trained  $F_{st}$ as a critic to guide the learning of sketch canvas-size selector via Reinforcement Learning ~\cite{kaelbling1996reinforcement} as rasterisation is a non-differentiable operation. There are two major objectives here: a) retain the original resolution accuracy and b) encourage the use of the lower resolution to improve the computational efficiency. 

\keypoint{Model:} 
Any sequential network from RNN \cite{collomosse2019livesketch} or Transformer \cite{lin2020sketch} families can be used to encode the vector-sketch $s_v$. To recap, sketch-vector ($s_v$) represents a sequence of points [$v_1$,$v_2$,...,$v_\text{T}$], where $v_t = (x_t, y_t, q^1_t, q^2_t, q^3_t ) \in \mathbb{R}^{T\times 5}$; T is the sequence length, ($x_t, y_t$) denotes the absolute coordinates in a normalised $c \times c$ canvas, while the last three represent pen-states~\cite{ha2018neural}. Furthermore, the complexity of canvas-size selector being dependent on sequence-length (T) varying across sketches, we use Douglas Peucker Algorithm~\cite{visvalingam1990douglas} to limit sequence-length across all sketch-samples at $\text{T}_{max}$ without losing visual representation. We feed $v_t$ at every time step to a simple 1-layer GRU network with $\mathbb{R}^{d_v}$ hidden states and take the final hidden state as the encoded latent representation ($f_{s^\text{T}} \in \mathbb{R}^{d_v}$). Next, we apply a linear layer ($\gamma$) to get $p(c|s_v) = \texttt{softmax}(W_\gamma f_{s^T} + b_\gamma)$, where $W_\gamma \in \mathbb{R}^{d_v\times K}$ and $b \in \mathbb{R}^K$, as shown in \cref{fig:network} (right). Given $p(c|s_{v}) \in \mathbb{R}^K$ over $K$ different canvas-sizes for every $s_{v}$, canvas-size $c$ can be sampled from categorical distribution as $c_{pred} \sim \texttt{categorical}([p(c_1|s_{v}), \cdots, p(c_K|s_{v})])$. Accordingly, $s_v$ is rasterised to $s_r$ at canvas-size $c_{pred}$, which is then fed to $F_{st}$. In particular, given a sketch $s_v$, canvas-size selector $\psi_C$ acting as a policy network, takes \textit{action} of selecting the canvas-size $c_{pred}$ from action space $C$, and $\psi_C$ is optimised over a \textit{reward} calculated by the SketchyNetV1 model ($F_{st}$) acting as the critic. 

\keypoint{Reward Design:} 
The objectives of our canvas-selector are to choose an optimal canvas-size such that: (i) accuracy is retained (ii) overall FLOPs is minimised. We therefore design our reward from two perspectives. Now as $F_{st}$ is fixed, we pre-compute the features of all gallery photos. During training, we only extract the feature of \textit{rasterised} sketch and calculate its corresponding rank using the pre-computed photo-features. 
Furthermore, taking one sketch we pre-compute FLOPs for every canvas-size in $C$.
From accuracy perspective, the aim is to select the optimal canvas-size $c_\text{opt}$ from $C$ for each $s_{v}$ such that the rendered $s_{r}$ can retrieve the matching photo ${p}$ at numerically lowest rank $r$ (best $r$=1). Following conventional norm of reward maximisation, we define accuracy reward ($R_\text{acc}$) as weighted (by hyper-parameters $\lambda_r$, $\lambda_\text{Tri}$) summation of inverse of the rank (r) and negative triplet loss (following Eq.\ref{equ:baseTrip}) as:
\vspace{-0.2cm}
{\small
\begin{equation}
\label{equ:rl_reward_acc}
\begin{aligned}
    R_\text{acc} = \lambda_{r} (\nicefrac{1}{r}) 
            + \lambda_\text{Tri}(-\mathcal{L}_\text{Tri}) 
\end{aligned}
\vspace{-0.25cm}
\end{equation}
}

\noindent From the compute perspective, the selection criterion should reward choosing a lower canvas-size. Moreover, higher performance being naturally inclined towards a higher-canvas-size thus increasing overall compute, we need an objective dedicated towards reducing computational cost. We thus propose a FLOPs constraint regularisation to guide the learning of canvas-selector as:
\vspace{-0.2cm}
{\small
\begin{equation}
\begin{aligned}
    \mathcal{L_\text{F}} = \frac{\sum_{j=1}^{K} (q_j \cdot \eta_j)}{q_{max} - q_{min}} 
\end{aligned}
\label{equ:flops}
\vspace{-0.1cm}
\end{equation}
}

\noindent where, $\eta_i = p(c_i|s_v)$ and $q_j$ is the pre-computed FLOPs value for the j-th canvas-size with maximum at $q_{max}$ and minimum at $q_{min}$.
Treating (-$\mathcal{L}_\text{F}$) as a reward from the compute perspective, we define our compute reward as $R_\text{comp} = (-\mathcal{L}_\text{F})$. Taking $\lambda_\text{F}$ as a balancing hyper-parameter, we combine both perspectives as our total reward:
\vspace{-0.4cm}
{\small
\begin{equation}
\vspace{-0.1cm}
    R_\text{Tot} = \lambda_\text{F} R_\text{comp} + (1-\lambda_\text{F} ) R_\text{acc}
\vspace{-0.1cm}
\label{equ:rl_reward}
\end{equation}
}

\keypoint{Optimisation Objective:} Finally, we train the canvas-size selector using popular Policy Gradient \cite{sutton2000policy} method which is both simple to implement and requires less hyper-parameter tuning than other state-of-the-art alternatives \cite{schulman2017proximal}. If $p(c|s_v^i)$ be the probability of selecting a specific canvas-size for $i^{th}$ sketch sample $s_v^i$ and the corresponding reward be $R_\text{Tot}^i$, the objective function for policy network (canvas-size selector) over a batch of size $B$ becomes:  

\vspace{-2mm}
{\small
\begin{equation}
    \mathcal{L}_{PG}(\theta) = - \frac{1}{B} \sum_{i=1}^{B} \log p(c|s_{v}^i) \cdot R_\text{Tot}^i
\vspace{-2mm}
\end{equation}
}

\vspace{-3.5mm}
\section{Experiments}
\label{sec:expt}
\vspace{-1mm}

{\setlength{\tabcolsep}{1.2pt}
\renewcommand{\arraystretch}{1}
\begin{table*}[t]
\tiny 
\centering
\caption{Quantitative Analysis on FG-SBIR. Best viewed when \textit{zoomed}.}
\vspace{-0.35cm}
\label{tab:quant}
\resizebox{\textwidth}{!}{
\begin{tabular}{lccccccccccccccc}
  \toprule
  \multicolumn{2}{c}{\multirow{2}{*}{Methods}} & Canvas-size & Params & \multicolumn{3}{c}{ShoeV2 \cite{yu2016sketch}} & \multicolumn{3}{c}{ChairV2 \cite{yu2016sketch}} & \multicolumn{3}{c}{Sketchy \cite{sangkloy2016sketchy}} & \multicolumn{3}{c}{FSCOCO \cite{chowdhury2022fs}} \\
  \cmidrule(lr){5-7}\cmidrule(lr){8-10}\cmidrule(lr){11-13}\cmidrule(lr){14-16}
  & & c$\times$c & (mil.) &  Top1 (\%) & Top10 (\%) &FLOPS (G) & Top1 (\%) & Top10 (\%) &FLOPS (G) & Top1 (\%) & Top10 (\%) &FLOPs (G) & Top1 (\%) & Top10 (\%) &FLOPS (G)\\
  \cmidrule(lr){1-4}\cmidrule(lr){5-7}\cmidrule(lr){8-10}\cmidrule(lr){11-13}\cmidrule(lr){14-16}
  \multirow{30}{*}{\rotatebox[origin=c]{90}{State-of-the-Arts}} 
  & \multirow{6}{*}{\rotatebox[origin=c]{90}{Triplet-SN~\cite{yu2016sketch}}} 
      & 32x32          & 8.75 & 09.77  \hlbad{($\downarrow$18.94)} & 24.82  \hlbad{($\downarrow$46.74)} & 0.083 & 15.61  \hlbad{($\downarrow$32.04)} & 29.86  \hlbad{($\downarrow$54.38)} & 0.083 &  4.23  \hlbad{($\downarrow$11.12)} & 12.58  \hlbad{($\downarrow$23.13)} & 0.083 & 0.76  \hlbad{($\downarrow$3.94)} &  6.22  \hlbad{($\downarrow$14.8)}  & 0.083 \\
    & & 64x64          & 8.75 & 17.63  \hlbad{($\downarrow$11.08)} & 44.96  \hlbad{($\downarrow$26.60)} & 0.338 & 30.25  \hlbad{($\downarrow$17.4)}  & 54.21  \hlbad{($\downarrow$30.03)} & 0.338 &  9.68  \hlbad{($\downarrow$5.67)}  & 22.45  \hlbad{($\downarrow$13.26)} & 0.338 & 1.85  \hlbad{($\downarrow$2.85)} &  7.33  \hlbad{($\downarrow$13.69)} & 0.338 \\
    & & 128x128        & 8.75 & 25.01  \hlbad{($\downarrow$3.70)}   & 62.35  \hlbad{($\downarrow$9.21)} & 1.397 & 40.61  \hlbad{($\downarrow$7.04)}  & 72.68  \hlbad{($\downarrow$11.56)} & 1.397 & 12.98  \hlbad{($\downarrow$2.37)}  & 29.63  \hlbad{($\downarrow$6.08)}  & 1.397 & 3.28  \hlbad{($\downarrow$1.42)} & 15.92  \hlbad{($\downarrow$5.1)}   & 1.397 \\
    & & 256x256        & 8.75 & 28.71            & 71.56            & 5.280 & 47.65            & 84.24            & 5.280 & 15.35            &35.71             & 5.280 &  4.7           & 21.02            & 5.280 \\
    & & SketchyNetV1   & 2.22 & 28.46 \hlgood{($\downarrow$0.25)} & 69.01 \hlgood{($\downarrow$2.55)} & 0.833  & 46.28 \hlgood{($\downarrow$0.37)} & 82.33 \hlgood{($\downarrow$1.91)} & 0.833 & 14.85 \hlgood{($\downarrow$0.5)}  & 34.68 \hlgood{($\downarrow$1.03)} & 0.833      & 4.22 \hlgood{($\downarrow$0.48)} & 18.33 \hlgood{($\downarrow$2.69)}  & 0.833 \\
    &  & \cellcolor{Gray}{SketchyNetV2}   & \cellcolor{Gray}2.27 & \cellcolor{Gray}27.89 \hlgood{($\downarrow$0.82)} & \cellcolor{Gray}68.76 \hlgood{($\downarrow$2.80)} & \cellcolor{Gray}0.264  & \cellcolor{Gray}45.98 \hlgood{($\downarrow$1.67)} & \cellcolor{Gray}80.14 \hlgood{($\downarrow$4.10)} & \cellcolor{Gray}0.294 & \cellcolor{Gray}14.21 \hlgood{($\downarrow$1.14)} & \cellcolor{Gray}34.16 \hlgood{($\downarrow$1.55)} & \cellcolor{Gray}0.321      & \cellcolor{Gray}3.98 \hlgood{($\downarrow$0.72)} & \cellcolor{Gray}17.64 \hlgood{($\downarrow$3.38)}  & \cellcolor{Gray}0.423 \\
       \cmidrule(lr){2-4}\cmidrule(lr){5-7}\cmidrule(lr){8-10}\cmidrule(lr){11-13}\cmidrule(lr){14-16}
    & \multirow{6}{*}{\rotatebox[origin=c]{90}{HOLEF-SN~\cite{song2017deep}} }
      & 32x32          & 9.31 & 10.84  \hlbad{($\downarrow$20.9)}  & 27.28  \hlbad{($\downarrow$48.5)}  & 0.096 & 18.72  \hlbad{($\downarrow$34.69)} & 31.68  \hlbad{($\downarrow$55.88)} & 0.096 &  6.01  \hlbad{($\downarrow$10.69)} & 14.12  \hlbad{($\downarrow$24.78)} & 0.096 & 0.86  \hlbad{($\downarrow$4.04)} & 6.93   \hlbad{($\downarrow$14.78)} & 0.096 \\
    & & 64x64          & 9.31 & 19.83  \hlbad{($\downarrow$11.91)} & 51.72  \hlbad{($\downarrow$24.06)} & 0.387 & 34.32  \hlbad{($\downarrow$19.09)} & 56.46  \hlbad{($\downarrow$31.1)}  & 0.387 &  9.92  \hlbad{($\downarrow$6.78)}  & 24.38  \hlbad{($\downarrow$14.52)} & 0.387 & 2.11  \hlbad{($\downarrow$2.79)} & 7.86   \hlbad{($\downarrow$13.85)} & 0.387 \\
    & & 128x128        & 9.31 & 26.78  \hlbad{($\downarrow$4.96)}  & 66.28  \hlbad{($\downarrow$9.5)}   & 1.545 & 45.74  \hlbad{($\downarrow$7.67)}  & 75.59  \hlbad{($\downarrow$11.97)} & 1.545 & 13.81  \hlbad{($\downarrow$2.89)}  & 33.68  \hlbad{($\downarrow$5.22)}  & 1.545 & 3.65  \hlbad{($\downarrow$1.25)} & 16.32  \hlbad{($\downarrow$5.39)}  & 1.545 \\
    & & 256x256        & 9.31 & 31.74             & 75.78            & 5.758 & 53.41            & 87.56            & 5.758 & 16.70            &38.90             & 5.758 & 4.9            & 21.71            & 5.758 \\
    & & SketchyNetV1   & 2.22 & 31.59 \hlgood{($\downarrow$0.15)} & 73.96 \hlgood{($\downarrow$1.82)} & 0.833 & 53.27 \hlgood{($\downarrow$0.14)} & 86.93 \hlgood{($\downarrow$0.63)}     & 0.833 & 16.25 \hlgood{($\downarrow$0.45)}  & 38.21 \hlgood{($\downarrow$0.69)}  & 0.833 & 4.56 \hlgood{($\downarrow$0.34)} & 19.92 \hlgood{($\downarrow$1.79)}  & 0.833 \\
    &   & \cellcolor{Gray}SketchyNetV2   & \cellcolor{Gray}2.27 & \cellcolor{Gray}30.86 \hlgood{($\downarrow$0.88)} & \cellcolor{Gray}72.56 \hlgood{($\downarrow$3.22)} & \cellcolor{Gray}0.259 & \cellcolor{Gray}52.74 \hlgood{($\downarrow$0.67)} & \cellcolor{Gray}84.02 \hlgood{($\downarrow$3.54)}     & \cellcolor{Gray}0.289 & \cellcolor{Gray}15.91 \hlgood{($\downarrow$0.79)}  & \cellcolor{Gray}37.88 \hlgood{($\downarrow$1.02)}  & \cellcolor{Gray}0.315 & \cellcolor{Gray}4.04 \hlgood{($\downarrow$0.86)} & \cellcolor{Gray}18.63 \hlgood{($\downarrow$3.08)}  & \cellcolor{Gray}0.417 \\
       \cmidrule(lr){2-4}\cmidrule(lr){5-7}\cmidrule(lr){8-10}\cmidrule(lr){11-13}\cmidrule(lr){14-16}
    & \multirow{6}{*}{\rotatebox[origin=c]{90}{Triplet-RL~\cite{bhunia2020sketch}}} 
      & 32x32          & 22.1 & 11.98  \hlbad{($\downarrow$22.12)} & 28.11  \hlbad{($\downarrow$50.71)} & 0.142 & 20.53  \hlbad{($\downarrow$36.01)} & 32.59  \hlbad{($\downarrow$57.02)} & 0.142 & 1.76  \hlbad{($\downarrow$2.94)} &  3.88  \hlbad{($\downarrow$6.47)} & 0.142 & -- & -- & -- \\
    & & 64x64          & 22.1 & 21.28  \hlbad{($\downarrow$12.82)} & 54.71  \hlbad{($\downarrow$24.11)} & 0.577 & 36.09  \hlbad{($\downarrow$20.45)} & 57.33  \hlbad{($\downarrow$32.28)} & 0.577 & 3.04  \hlbad{($\downarrow$1.66)} &  6.79  \hlbad{($\downarrow$3.56)} & 0.577 & -- & -- & -- \\
    & & 128x128        & 22.1 & 28.83  \hlbad{($\downarrow$5.27)}  & 67.84  \hlbad{($\downarrow$10.98)} & 2.299 & 48.38  \hlbad{($\downarrow$8.16)}  & 77.25  \hlbad{($\downarrow$12.36)} & 2.299 & 4.12  \hlbad{($\downarrow$0.58)} &  9.05  \hlbad{($\downarrow$1.30)} & 2.299 & -- & -- & -- \\
    & & 256x256        & 22.1 & 34.10             & 78.82            & 6.041 & 56.54            & 89.61            & 6.041 & 4.70            & 10.35            & 6.041 & -- & -- & -- \\
    & & SketchyNetV1   & 2.22 & 33.88  \hlgood{($\downarrow$0.22)} & 77.15 \hlgood{($\downarrow$1.67)} & 0.833 & 55.92 \hlgood{($\downarrow$0.62)} & 88.20 \hlgood{($\downarrow$1.41)} & 0.833 & 4.59 \hlgood{($\downarrow$0.11)} & 10.21 \hlgood{($\downarrow$0.14)} & 0.833      & -- & -- & -- \\
    &   & \cellcolor{Gray}SketchyNetV2   & \cellcolor{Gray}2.27 & \cellcolor{Gray}33.26  \hlgood{($\downarrow$0.84)} & \cellcolor{Gray}76.84 \hlgood{($\downarrow$1.98)} & \cellcolor{Gray}0.255 & \cellcolor{Gray}55.14 \hlgood{($\downarrow$1.40)} & \cellcolor{Gray}87.31 \hlgood{($\downarrow$2.30)} & \cellcolor{Gray}0.285 & \cellcolor{Gray}4.51 \hlgood{($\downarrow$0.19)} &  \cellcolor{Gray}9.89 \hlgood{($\downarrow$0.46)} & \cellcolor{Gray}0.308      & \cellcolor{Gray}-- &\cellcolor{Gray} -- &\cellcolor{Gray} -- \\
       \cmidrule(lr){2-4}\cmidrule(lr){5-7}\cmidrule(lr){8-10}\cmidrule(lr){11-13}\cmidrule(lr){14-16}
    & \multirow{6}{*}{\rotatebox[origin=c]{90}{StyleVAE~\cite{sain2021stylemeup}}}
      & 32x32        & 25.37 & 12.68  \hlbad{($\downarrow$23.79)} & 28.69  \hlbad{($\downarrow$53.14)} & 0.125 & 22.48 \hlbad{($\downarrow$40.38)} & 32.10 \hlbad{($\downarrow$59.04)} & 0.125 & 06.92 \hlbad{($\downarrow$12.7)} & 16.25 \hlbad{($\downarrow$29.53)}  & 0.125 & -- & -- & -- \\
    & & 64x64        & 25.37 & 22.93  \hlbad{($\downarrow$13.54)} & 57.71  \hlbad{($\downarrow$24.12)} & 0.508 & 40.26 \hlbad{($\downarrow$22.6)}  & 58.23 \hlbad{($\downarrow$32.91)} & 0.508 & 12.46 \hlbad{($\downarrow$7.16)} & 29.62 \hlbad{($\downarrow$16.16)}  & 0.508 & -- & -- & -- \\
    & & 128x128      & 25.37 & 30.91  \hlbad{($\downarrow$5.56)}  & 70.06  \hlbad{($\downarrow$11.77)} & 2.024 & 53.88 \hlbad{($\downarrow$8.98)}  & 78.64 \hlbad{($\downarrow$12.5)}  & 2.024 & 17.15 \hlbad{($\downarrow$2.47)} & 38.84 \hlbad{($\downarrow$6.94)}   & 2.024 & -- & -- & -- \\
    & & 256x256      & 25.37 & 36.47                              & 81.83                              & 5.642  & 62.86                             & 91.14                             & 5.642  & 19.62                             & 45.78                             & 5.642  & -- & -- & -- \\
    & & SketchyNetV1 &  2.22 & 36.11 \hlgood{($\downarrow$0.36)}  & 79.63 \hlgood{($\downarrow$2.20)}  & 0.833 & 62.23 \hlgood{($\downarrow$0.63)} & 88.95 \hlgood{($\downarrow$2.19)} & 0.833 & 19.48 \hlgood{($\downarrow$0.12)} & 44.02 \hlgood{($\downarrow$1.76)} & 0.833       & -- & -- & -- \\
    &  & \cellcolor{Gray}SketchyNetV2 & \cellcolor{Gray}2.27 & \cellcolor{Gray}35.88 \hlgood{($\downarrow$0.59)}  & \cellcolor{Gray}78.46 \hlgood{($\downarrow$3.37)}  & \cellcolor{Gray}0.251 & \cellcolor{Gray}61.95 \hlgood{($\downarrow$0.91)} & \cellcolor{Gray}87.68 \hlgood{($\downarrow$3.46)} & \cellcolor{Gray}0.281 & \cellcolor{Gray}19.10 \hlgood{($\downarrow$0.50)} & \cellcolor{Gray}43.57 \hlgood{($\downarrow$2.21)} & \cellcolor{Gray}0.304       & \cellcolor{Gray}-- & \cellcolor{Gray}-- & \cellcolor{Gray}-- \\
       \cmidrule(lr){2-4}\cmidrule(lr){5-7}\cmidrule(lr){8-10}\cmidrule(lr){11-13}\cmidrule(lr){14-16}
    & \multirow{6}{*}{\rotatebox[origin=c]{90}{Partial-OT~\cite{chowdhury2022partially}}} 
      & 32x32          & 22.1 & -- & -- & -- & -- & -- & -- & -- & -- & -- &  6.11  \hlbad{($\downarrow$18.05)} & 24.18  \hlbad{($\downarrow$29.74)} & 0.196 \\
    & & 64x64          & 22.1 & -- & -- & -- & -- & -- & -- & -- & -- & -- & 12.32  \hlbad{($\downarrow$11.84)} & 37.65  \hlbad{($\downarrow$16.27)} & 0.687 \\
    & & 128x128        & 22.1 & -- & -- & -- & -- & -- & -- & -- & -- & -- & 19.61  \hlbad{($\downarrow$4.55)}  & 45.12  \hlbad{($\downarrow$8.8)}   & 2.876 \\
    & & 256x256        & 22.1 & -- & -- & -- & -- & -- & -- & -- & -- & -- & 24.16                              & 53.92                              & 6.512 \\
    & & SketchyNetV1   & 2.22 & -- & -- & -- & -- & -- & -- & -- & -- & -- & 23.89  \hlgood{($\downarrow$0.27)} & 51.29 \hlgood{($\downarrow$2.63)}  & 0.833 \\
    &   & \cellcolor{Gray}SketchyNetV2   & \cellcolor{Gray}2.27 & \cellcolor{Gray}-- &\cellcolor{Gray} -- & \cellcolor{Gray}-- &\cellcolor{Gray} -- &\cellcolor{Gray} -- &\cellcolor{Gray} -- &\cellcolor{Gray} -- &\cellcolor{Gray} -- &\cellcolor{Gray} -- & \cellcolor{Gray}23.77  \hlgood{($\downarrow$0.39)} & \cellcolor{Gray}50.68 \hlgood{($\downarrow$3.24)}  & \cellcolor{Gray}0.325 \\
    \cmidrule(lr){1-4}\cmidrule(lr){5-7}\cmidrule(lr){8-10}\cmidrule(lr){11-13}\cmidrule(lr){14-16}
    \multirow{11}{*}{\rotatebox[origin=c]{90}{Baselines}} 
    & B-BFR        & 224x224 & 4.61 & 24.32 & 61.23 & 2.602 & 43.39 & 76.33 & 2.602 & 10.47 & 24.34 & 2.602 & 2.15 & 11.61 & 2.602 \\
    & B-DRS        & Dynamic & 9.36 & 25.12 & 68.84 & 5.493 & 44.78 & 78.14 & 6.187 & 11.87 & 26.58 & 6.245 & 3.86 & 17.63 & 6.291 \\
    & B-Crop       & Dynamic & 4.92 & 20.32 & 52.12 & 6.562 & 35.07 & 57.68 & 6.597 & 08.27 & 20.19 & 6.629 & 1.98 & 09.67 & 6.688 \\
       \cmidrule(lr){2-4}\cmidrule(lr){5-7}\cmidrule(lr){8-10}\cmidrule(lr){11-13}\cmidrule(lr){14-16}
    & B-Regress \cite{romero2015fitnets}    & Dynamic & 2.27 & 22.82 & 61.33 & 0.261 & 38.42 & 64.65 & 0.291 & 08.69 & 21.98 & 0.323 & 2.07 & 10.15 & 0.375 \\
    & B-AKD \cite{zagoruyko2017KDattention} & Dynamic & 2.27 & 23.67 & 64.26 & 0.268 & 38.61 & 71.88 & 0.295 & 12.67 & 27.71 & 0.331 & 2.91 & 14.01 & 0.392 \\
    & B-PKT \cite{passalis2018learning}     & Dynamic & 2.27 & 24.31 & 65.91 & 0.259 & 42.68 & 73.66 & 0.288 & 12.27 & 28.31 & 0.317 & 3.45 & 16.22 & 0.372 \\
       \cmidrule(lr){2-4}\cmidrule(lr){5-7}\cmidrule(lr){8-10}\cmidrule(lr){11-13}\cmidrule(lr){14-16}
    & B-VGG16-SN    & 256x256      & 14.71 & 33.03                              & 78.51                              & 40.18 & 52.16                              & 84.08                              & 40.18 & 18.61                             & 41.25                             & 40.18 & 5.16                             & 23.62                              & 40.18 \\
    & B-ThiNet~\cite{luo2017thinet}     & 256x256      &  8.32 &  7.51 \hlbad{($\downarrow$25.52)}  & 22.34  \hlbad{($\downarrow$56.17)} & 9.342 & 12.31 \hlbad{($\downarrow$39.85)}  & 26.63 \hlbad{($\downarrow$57.45)}  & 9.342 & 3.16  \hlbad{($\downarrow$15.45)} & 12.73 \hlbad{($\downarrow$28.52)} & 9.342 & 0.65  \hlbad{($\downarrow$4.51)} &  5.35 \hlbad{($\downarrow$18.27)}  & 9.342 \\
    & B-Prune~\cite{molchanov2016pruning} & 256x256      &  6.17 &  3.92 \hlbad{($\downarrow$29.11)}  & 17.32  \hlbad{($\downarrow$61.19)} & 8.138 &  7.98 \hlbad{($\downarrow$44.18)}  & 19.82 \hlbad{($\downarrow$64.26)}  & 8.138 & 1.45  \hlbad{($\downarrow$17.16)} &  8.22 \hlbad{($\downarrow$33.03)} & 8.138 & 0.53  \hlbad{($\downarrow$4.63)} &  4.97 \hlbad{($\downarrow$18.65)}  & 8.138 \\
    & \cellcolor{Gray}B-VGG16-SN    & \cellcolor{Gray}SketchyNetV2 & \cellcolor{Gray} 2.27 &\cellcolor{Gray} 32.77 \hlgood{($\downarrow$0.26)} & \cellcolor{Gray}78.21 \hlgood{($\downarrow$0.3)}    & \cellcolor{Gray}0.254 & \cellcolor{Gray}50.75 \hlgood{($\downarrow$1.41)}  & \cellcolor{Gray}82.21 \hlgood{($\downarrow$1.87)}  & \cellcolor{Gray}0.283 & \cellcolor{Gray}16.55 \hlgood{($\downarrow$2.06)} & \cellcolor{Gray}38.66 \hlgood{($\downarrow$2.59)} & \cellcolor{Gray}0.332 & \cellcolor{Gray}4.69 \hlgood{($\downarrow$0.47)} &\cellcolor{Gray} 21.76 \hlgood{($\downarrow$1.86)}  & \cellcolor{Gray}0.407 \\
    
    \bottomrule
    \end{tabular}
}
\vspace{-4mm}
\end{table*}
}

\noindent\textbf{Datasets:}
We evaluated on the following publicly available datasets having fine-grained sketch-photo association. \textbf{QMUL-ShoeV2}~\cite{yu2016sketch} and \textbf{QMUL-ChairV2}~\cite{yu2016sketch} contains $6730$/$2000$ and $1800$/$400$ sketches/photos respectively. Keeping $679$/$200$ ($525$/$100$) sketches/photos for evaluation from ShoeV2 (ChairV2) respectively we use the rest for training. \textbf{Sketchy} \cite{sangkloy2016sketchy} contains $125$ categories with $100$ photos each, having around 5 sketches per photo. While, training uses a standard train-test split \cite{sangkloy2016sketchy} of $9$:$1$, during inference we construct a challenging gallery using photos across one category for retrieval. For scene-level FG-SBIR evaluation we use \textbf{FS-COCO}~\cite{chowdhury2022fs} which includes $10$,$000$ unique sketch-photo pairs with a $7$:$3$ train/test split.

\vspace{+1mm}
\noindent\textbf{Implementation Details:}
Considering a standard large FG-SBIR network as a teacher, our proposed student uses a cheaper backbone feature extractor MobileNetV2~\cite{sandler2018mobilenetv2}, a standard for mobile end-devices. Margin is set to $m = 0.2$ for triplet loss (\cref{sec:Pilot,sec:student_model}). We use an Adam optimiser at a learning rate of 0.0001 and a batchsize of 16 for 200 epochs. For practicality, we chose discrete canvas-sizes ($C$) instead of a continuous action space, where $C$ = $\{32\times32$, $64\times64$, $128\times128$, $256\times256\}$. Canvas-size selector ($\psi_C$) is modelled using a simple GRU network, with embedding size 128, and trained via reinforcement learning with a learning rate of 0.0001 and a batchsize of 32 for 500 epochs, keeping $\text{T}_{max}$=100 for vector-sketches.
For the action space, we take varying canvas-sizes from $C$ and set $\beta = 1$ in Eq.\ref{equ:loss_rkd}.  $\lambda$, $\lambda_{r}$, $\lambda_\text{Tri}$ and $\lambda_\text{F}$ are empirically set to 0.5, 0.4, 0.48 and 0.35 (Fig.\red{5}) respectively. 
{While training $\psi_C$, sketches are rendered at different completion (30\%, 35\%, ... 100\%) levels to simulate varying levels of \textit{abstraction}, ensuring a \textit{sketch-abstraction-aware} canvas-selector (\textit{e.g.}, a \textit{smaller} canvas for \textit{less detailed} sketch).}

\keypoint{Evaluation Metric:} 
\textit{Accuracy} for FG-SBIR is the percentage of sketches where true-match photos are ranked in the Top-q (q = 1,10) lists. \textit{Computational efficiency} is measured via parameter-count and average FLOPs -- lower is better. For average FLOPs in \Cref{tab:quant}, we take the combined sum of FLOPs per sample \cite{chen2022dynamic} from (i) canvas-size selector and (ii) retrieval model, using the predicted resolution, and average across the validation set.

\subsection{Competitors}
\vspace{-3mm}
\label{sec:competitors}

\noindent We evaluate against State-of-the-Arts (SoTAs) and a few relevant curated baselines (\textbf{B-}). \textbf{(i)~SoTAs:} {Unlike \textit{Triplet-SN}~\cite{yu2016sketch}, \textit{HOLEF-SN}~\cite{song2017deep}, \textit{Triplet-RL}~\cite{bhunia2020sketch} (we report \textit{full}-sketch evaluation) and \textit{StyleVAE} \cite{sain2021stylemeup}; \textit{Partial-OT}~\cite{chowdhury2022partially} specifically caters to scene-level FG-SBIR. For fairness, we vary input-resolution of sketches during inference (fixed at their original $256\times256$ setup during training). We also adapt them as teachers, to judge our accuracy retainment after model-compression (\textit{SketchyNetV1}) using full-res sketch and, further optimising canvas-size with canvas-selector (\textit{SketchyNetV2}).} \textbf{(ii) KD baselines:} Here we re-purpose existing works in a cross-modal setting, keeping our canvas-selector same, with \textit{Triplet-SN} as the teacher. \textit{B-Regress} -- regresses features from teacher and student for both sketches and images, aligning their dimensions with an extra transformation layer \cite{romero2015fitnets}, using a $l_2$ regression loss. Following \cite{passalis2018learning} \textit{B-PKT} calculates the conditional probability density for any pair of points in the teacher's embedding space~\cite{passalis2018learning}, which models the probability of any two samples being close together. Sampling 'N' instances, it establishes a probability distribution for pairwise interactions within that space. Simultaneously, in the student's embedding space, a comparable distribution is derived, and the model minimises the divergence between them using a KL-divergence loss. \textit{B-AKD} follows \cite{zagoruyko2017KDattention} in using spatial attention maps for knowledge transfer using similar dimensional matching like \textit{B-Regress}. \noindent\textbf{(iii) Resizing strategies:} Keeping the same KD-paradigm with \textit{Triplet-SN}-teacher, we vary the input-size resizing module following existing works. \textit{B-BFR} follows \cite{talebi2021learning} in designing a bilinear feature resizer amidst the intermediate layers of a CNN network, to output a down-scaled image. We choose its resized resolution of 224$\times$224 (lowest) for subsequent evaluation. \textit{B-DRS} follows \cite{chen2022dynamic} in training a convolutional module that takes an image in full-resolution, calculates a probability vector on available canvas-sizes, and transforms it into binary decisions, indicating the scale factor of selection. \textit{B-Crop} follows \cite{li2020learning} in meta-learning an image-cropping module that crops the input image to a target resolution for low compute evaluation. \noindent\textbf{(iv) Pruning alternatives:} {Here we adapt two VGG16-backboned methods off-the-shelf from \textit{network-pruning family} among other efficiency paradigms ~\cite{gupta2015deep, courbariaux2015binaryconnect, talebi2021learning}, in FG-SBIR setup -- \textit{B-ThiNet}~\cite{luo2017thinet} and \textit{B-Prune}~\cite{molchanov2016pruning}. For comparative reference we report performance of a VGG16-backboned \textit{Triplet-SN} network (\textit{B-VGG16-SN}) and its \textit{SketchyNetV2} variant.}

\vspace{-2mm}
\subsection{Performance and Compute Analysis:}
\vspace{-2mm}

\keypoint{Analysis on Accuracy:}
\Cref{tab:quant} reports the quantitative evaluation of our model against others. \textit{Triplet-SN}~\cite{yu2016sketch} and \textit{HOLEF-SN}~\cite{song2017deep} perform lower owing to weak backbones of Sketch-A-Net \cite{Yu2015SketchaNetTB}. Enhanced by its RL-optimised reward function \textit{Triplet-RL}\cite{bhunia2020sketch} scores higher (2.36\%$\uparrow$ Top1 on ShoeV2) but fails to surpass \textit{StyleVAE} \cite{sain2021stylemeup}, due to its delicate meta-learned disentanglement module addressing style-diversity. Being trained on region-wise sketch-photo associativity, {\textit{Partial-OT}~\cite{chowdhury2022partially} outperforms all others significantly on scene-level sketches (FS-COCO).} We also notice a gradual drop in accuracy of SoTAs, when subjected to decreasing resolution of input-sketches suggesting that some methods are inherently accustomed to handling scale-shifts better than others, but are inferior to our dynamic approach (\textit{SketchyNetV2}). Importantly, when most SoTAs are employed as teachers, their respective students (\textit{SketchyNetV1} variants) are seen to retain their respective original accuracy with a marginal drop of $\approx0.5-1\%$ overall, thanks to our efficient KD paradigm. Having different teacher-student embedding spaces both \textit{B-Regress} and \textit{B-AKD}'s paradigm of directly regressing on features proves incompatible to a cross-modal retrieval setting, dropping accuracy (compared to \textit{T-Triplet-SN}), whereas lacking cross-modal discrimination \textit{B-PKT} performs lower than ours, proving our superiority in retaining accuracy. 

\vspace{-2mm}
\begin{figure}[h]
    \centering
    \includegraphics[width=\linewidth]{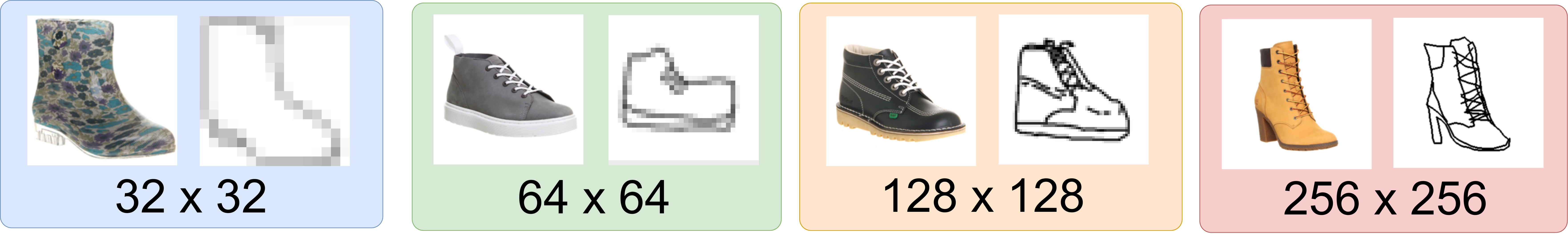}
    \vspace{-6mm}
    \caption{Exemplar sketches at their optimal canvas-size}
    \label{fig:optimal_canvas}
    \vspace{-4mm}
\end{figure}

\vspace{+1mm}
\noindent \textbf{Analysis on Compute:}
Although FLOPs \cite{chen2022dynamic} decrease (\Cref{tab:quant}) with a reduction in input resolution across all SoTAs, it comes at a severe cost of accuracy ($\approx$21\% average drop from 256$\times$256 to 32$\times$32 on ShoeV2). This is mainly because \textit{ideal} canvas-size for \textit{optimal} retrieval varies across samples (\cref{fig:pilot_plot} right) -- thus subjecting all samples to a fixed canvas-size proves detrimental for accuracy. However, a positive Acc@1 score of $\approx$11\% even at a low canvas-size of 32$\times$32 for every SoTA shows that some sketches are indeed recognisable if the model is trained properly. This is verified further by \textit{SketchyNetV2} variant of respective SoTAs, that secures the lowest FLOPs per SoTA, despite maintaining accuracy at par with its SoTA-counterpart -- thanks to our RL-guided canvas-selector, which can allot an optimal canvas-size per sketch towards better retrieval (\cref{fig:optimal_canvas}). Contrarily, \textit{B-BFR} fails to lower FLOPs as it fixes a 224x224 input resolution for a satisfactory Acc@1 compared to \textit{Triplet-SN}. \textit{B-Crop} fairs better, however sketch being sparse, cropped sketches leads to confused representations, lowering accuracy. Despite coming closest in accuracy to ours on \textit{Triplet-SN}, \textit{B-DRS} lags behind considerably on compute, thanks to our canvas-selector working in vector space (inputs $s_v$) and eliminating costly convolutional operations, unlike others using full-resolution sketch-raster as input. {Being untrained to handle \textit{sparse-nature} of sketches during such \textit{high} compression, performance of both \textit{B-ThiNet} and \textit{B-Prune} collapses (\vs \textit{B-VGG16-SN}) unlike its \textit{SketchyNetV2}.} Notably, the significant drop in parameter-count (4 - 10 times) against SoTAs, comes from choosing lightweight MobileNetV2~\cite{sandler2018mobilenetv2} as a student coupled with our canvas-size selector in every \textit{SketchyNetV2} variant. Compared to ShoeV2, FLOPs increase slightly for ChairV2, Sketchy and FS-COCO, owing to higher object (ChairV2) or scene-level (FS-COCO) details, and lower structural correspondence with photos (Sketchy), thus training a poorer retrieval model which in turn acts as a poor critic for our canvas-selector. Our canvas-selector alone takes a negligible 51.84K parameters with 0.02G FLOPs.

\vspace{-1mm}
\subsection{Ablation Study}
\label{sec:ablation}
\vspace{-1mm}

\noindent As our teacher ($F_T$) can be any pre-trained FG-SBIR network, we limit our ablation to justifying design choices of our student model and canvas-size selector. We use \textit{B-VGG16-SN} as a reference $F_T$ -- \textit{Type-$\emptyset$} in \Cref{tab:abla}.

\begin{figure}[h]
\centering
\vspace{-2mm}
\includegraphics[width=0.9\linewidth]{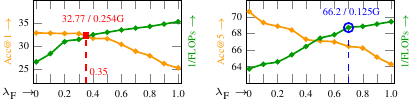}
\vspace{-5mm}
\caption{Impact of varying $\lambda_\text{F}$ on Acc@1 and Acc@5}
\label{fig:lambda-flops}
\vspace{-4mm}
\end{figure}

\keypoint{Training Canvas-Size selector $\psi_C$:} To justify loss objectives of $\psi_C$ we evaluate on ShoeV2, eliminating $r^\text{-1}$, $\mathcal{L}^R_\text{Tri}$, $\mathcal{L}_\text{F}$ one by one from Eq.~\ref{equ:rl_reward} (\Cref{tab:abla}). While eliminating accuracy rewards, (\textit{w/o rank$^{-1}$}, \textit{w/o $\mathcal{L}^R_\text{Tri}$}) increases constraint on lowering FLOPs, dropping accuracy, ignoring $\mathcal{L}_\text{F}$ (\textit{w/o $\mathcal{L}_\text{F}$}) gains accuracy but costs high FLOPs. We thus optimally balance between the two using $\lambda_\text{F}$ (\cref{fig:lambda-flops} left). 
One can also choose to train for \textit{faster} retrieval (lower FLOPs), given a decent Acc@5 (\cref{fig:lambda-flops} right).

\vspace{-1mm}
\keypoint{Architectural insights:} To judge the efficiency of our student's backbone, we explore a few cheaper backbone-networks (\textit{Types IV, V} in \Cref{tab:abla}). Evidently, MobileNetV2 (Ours) balances optimally between compute and accuracy. As an option to vector input for canvas-selector, we re-design it with a simple seven-layer CNN module following \cite{zhu2021dynamic} to use the \textit{full-resolution} sketch-image as its input, and select a target canvas-size to downscale it for retrieval. With extra computational cost coming from spatial domain, total FLOPs incurred is much higher (\textit{VI}).
Next we compare efficacy of sketch as input to $\psi_C$ with respect to types of vector format -- offset-coordinate (\textit{VI}) \cite{ha2018neural} \vs absolute ones (\textit{Ours}). Turns out the former is better for encoding. Comparing sketch encoders among LSTM (\textit{VIII}), GRU ({Ours}) and Transformer (\textit{IX}), showed ours to be optimum empirically (\Cref{tab:abla}). Furthermore, prepending recent pre-processing modules like \textit{stroke-subset selector} of \cite{bhunia2022sketching} to our canvas-selector, yields 42.95\% (\vs 43.7\%~\cite{bhunia2022sketching}) Acc@1 ShoeV2, proving its off-the-shelf compatibility.

\begin{table}[h]
    \vspace{-2mm}
    \setlength{\tabcolsep}{4pt}
    \renewcommand{\arraystretch}{0.6}
    \centering
    \scriptsize
    \caption{Ablative studies (accuracy on ShoeV2)}
    \vspace{-0.3cm}
    \begin{tabular}{clcccc}
        \toprule
        \multicolumn{2}{c}{\multirow{2}{*}{Type}} & FLOPs & Params & \multicolumn{2}{c}{ShoeV2}\\
         \cmidrule(lr){3-4}\cmidrule(lr){5-6}
        & &  (G) & (mil.) & Top1 (\%) & Top10 (\%)\\
        \cmidrule(lr){1-2}\cmidrule(lr){3-4}\cmidrule(lr){5-6}
        $\emptyset$ & B-VGG16-SN                        & 40.18  & 14.71 & 33.03 & 78.51 \\
        \cmidrule(lr){1-2}\cmidrule(lr){3-4}\cmidrule(lr){5-6}
        I & w/o rank$^\text{-1}$-reward         & 0.173 & \multirow{3}{*}{2.27} & 25.31 & 64.36 \\
        II & w/o $\mathcal{L}^R_\text{Tri}$-reward      & 0.182 &  & 27.78 & 69.32\\
        III & w/o $\mathcal{L}_\text{F}$-reward         & 0.833 &  & 32.91 & 78.39 \\
        \cmidrule(lr){1-2}\cmidrule(lr){3-4}\cmidrule(lr){5-6}
        IV & ResNet18 backbone                          & 1.451  & 11.18 & 26.72 & 68.18 \\
        V & EfficientNet backbone                       & 0.459  & 4.01 & 29.47 & 72.04 \\
        \cmidrule(lr){1-2}\cmidrule(lr){3-4}\cmidrule(lr){5-6}
        VI & Image-based $\psi_C$                       & 5.730  & 7.62 & 32.21 & 77.68 \\
        VII & Offset $s_v$                              & 0.255  & 2.27 & 32.61 & 78.40 \\
        VIII & Decoder-LSTM                             & 0.268  & 2.25 & 31.41 & 77.32 \\
        IX & Decoder-Tf                                 & 0.314  & 2.34 & 31.98 & 78.02 \\
        \cmidrule(lr){1-2}\cmidrule(lr){3-4}\cmidrule(lr){5-6}
        \rowcolor{YellowGreen!40}
        \multicolumn{2}{c}{SketchyNetV2}                 & 0.254  & 2.27 & 32.77 & 78.21 \\
        \bottomrule
    \end{tabular}
    \label{tab:abla}
    \vspace{-3mm}
\end{table}

\vspace{-1mm}
\keypoint{Generalisability of canvas selector:} To judge if the optimal canvas-size selected is uniform \textit{irrespective} of the training method, we first train \textit{four} canvas-selectors using four SoTAs of \Cref{tab:quant}. Evaluating them on ShoeV2~\cite{yu2016sketch}, 89.21\% of test-set sketches in ShoeV2 to obtain the \textit{same} optimal canvas size across \textit{four} different models, trained on four \textit{different} methods, which further proves that our canvas-selector can work equally well as a pre-processing module for any new model. Furthermore, plugging a canvas-selector trained with SketchyNetV1 (critic) from B-VGG16-SN with a different SketchyNetV1 from StyleVAE \cite{sain2021stylemeup} during inference achieves comparable performance ($35.69\%$ Top-1) to a canvas-selector trained on StyleVAE-SketchyNetV1 ($35.88\%$ Top-1) on ShoeV2. Importantly, $95.64\%$ of test-set sketches yield the same optimal canvas-size across both models, confirming the \textit{cross-model} generalisability of our canvas selector. \textit{(ii)} On training canvas-selector with teacher $F_T$ (for Triplet-SN) as critic, yields \textit{same} optimal canvas-sizes as those for student $F_S$, for $94.21\%$ of test-set sketches in ShoeV2, while incurring comparable FLOPs reduction (0.268G) and accuracy ($27.91\%$ Top1) against $F_S$ as critic.

\noindent \textbf{Comparing Inference times:} We evaluated on a 12 GB Nvidia RTX 2080 Ti GPU, on ShoeV2, for inference time of Triplet-SN, HOLEF-SN, Triplet-RL, and StyleVAE (SOTAs) on full-sketch to obtain 37, 38, 36 and 42 ms, against 10, 11, 11, and 12 ms for their SketchyNetV2 variants respectively. For \textit{practical} validation, we evaluated onnx-INT8 version of Triplet-SN \cite{yu2016sketch} and its SketchyNetV2 variant (ours) after post-training quantisation, on an iPhone13 via the \href{https://ai-benchmark.com/}{AI-benchmark app} for ShoeV2~\cite{yu2016sketch} to obtain 51ms and 18ms respectively, which shows our efficiency.

\vspace{+1mm}
\noindent \textbf{Retraining SoTA on mixed resolution sketches?:}
To explore if retraining SoTAs on sketches with mixed resolutions, we re-train Triple-SN \cite{yu2016sketch} on rendered mixed-resolution sketches of ShoeV2. Poor performance in \Cref{tab:retrain} at lower resolutions similar to \Cref{tab:quant} validates our KD approach. 
The key lies in that the sketch-feature from teacher used for student-guidance ($f_s^T$) is always extracted from \textit{full-resolution} sketch while that extracted by student is at varying resolutions, imparting scale invariance, which is ideally suited to our \textit{cross-modal retrieval} setup.

{\setlength{\tabcolsep}{2pt}
\vspace{-2mm}
\begin{table}[h]
    \scriptsize
    \centering
    \caption{Evaluating SoTAs trained on mixed-resolution sketches}
    \vspace{-2mm}
    \begin{tabular}{crrcrrc}
    \toprule
    \multirow{2}{*}{\makecell[c]{Canvas\\Size}} & \multicolumn{3}{c}{Triplet-SN} & \multicolumn{3}{c}{Triplet-RL} \\
    \cmidrule(lr){2-4} \cmidrule(l){5-7}
                      & Top1 (\%)  & Top10 (\%) & FLOPs (G) & Top1 (\%)  & Top10 (\%) & FLOPs (G)\\
    \cmidrule(r){1-1} \cmidrule(lr){2-4} \cmidrule(l){5-7}
      32$\times$32   & 10.91 & 26.65 & 0.083 & 12.97  & 31.08 & 0.142 \\
      64$\times$64   & 19.04 & 48.12 & 0.338 & 22.96  & 56.23 & 0.577 \\
     128$\times$128  & 26.07 & 63.87 & 1.397 & 30.04  & 72.95 & 2.299 \\
     256$\times$256  & 28.74 & 71.68 & 5.280 & 34.21  & 77.24 & 6.041 \\
    \bottomrule
    \end{tabular}
    \label{tab:retrain}
    \vspace{-4mm}
\end{table}
}

\subsection{Further Insights}
\vspace{-2mm}

\vspace{+1mm}
\noindent \textbf{Cross-dataset Generalisation:}
To judge the generalisation potential of our canvas-size selector across datasets we train it on ShoeV2, but plug it to a retrieval model (SketchyNetV1) trained on ChairV2, and evaluate on ChairV2. Against the standard result (\Cref{tab:quant}) of training both modules (SketchyNetV2) on ChairV2 (0.285G, 55.32\% Acc@1), this setup retains accuracy at 53.14\% with slight rise in FLOPs (0.314G), thus confirming its potential.

\vspace{+1mm}
\noindent \textbf{\textit{Faster} Sketch-recognition:} Exploring potential of our canvas-size selector, we prepend it to a standard \textit{sketch-recognition} pipeline \cite{Yu2015SketchaNetTB} to optimise the canvas-size of query sketch, and reduce overall compute. Unlike our $\psi_C$, it uses negative of cross-entropy loss as accuracy reward ($R_\text{acc}$), keeping rest same. Compared to existing Sketch-a-Net \cite{Yu2015SketchaNetTB} incurring 5.28G FLOPs at 68.71\% accuracy on Quickdraw \cite{ha2018neural}, our canvas-selector fitted paradigm drops FLOPs to 0.261G while retaining accuracy at 68.14\%. This shows for the first time, that sparsity of sketches can be handled, in various downstream tasks via our $\psi_C$.

\begin{table}[t]
\setlength{\tabcolsep}{6pt}
\renewcommand{\arraystretch}{0.8}
\scriptsize
\centering
\caption{Quantitative Evaluation on Categorical SBIR}
\vspace{-2mm}
\label{tab:quant_sBIR}
\begin{tabular}{llcccc}
\toprule
\multicolumn{2}{c}{\multirow{2}{*}{Methods}}  & FLOPs & Params & \multicolumn{2}{c}{Sketchy(ext)} \\
\cmidrule(lr){3-4}\cmidrule(lr){5-6}
& & (G) & (mil.)  & mAP@all & P@200 \\
\cmidrule(lr){1-2}\cmidrule(lr){3-4}\cmidrule(lr){5-6}
\multirow{4}{*}{\rotatebox[origin=c]{90}{SoTA}} 
& B-SBIR-SN                             &  40.18 & 14.71  & 0.715 & 0.861\\
& DSH ~\cite{liu2017deep}             &  22.14 & 16.32  & 0.711 & 0.858 \\ 
& B-D2S~\cite{dey2019doodle}            &  17.75  & 136.36  & 0.810 & 0.894\\
& StyleVAE ~\cite{sain2021stylemeup}    &  8.116 & 25.37  & 0.905 & 0.927 \\
\cmidrule(lr){1-2}\cmidrule(lr){3-4}\cmidrule(lr){5-6}
\multirow{4}{*}{\rotatebox[origin=c]{90}{T-SoTA}} 
& T-SBIR-SN                     &  0.254 & \multirow{4}{*}{2.27} & 0.702 & 0.810\\
& T-DSH                         &  0.289 &  &  0.704 & 0.823 \\
& T-D2S                         &  0.268 &  &  0.798 & 0.845\\
& T-StyleVAE                    &  0.251 &  &  0.886 & 0.897\\
\bottomrule
\end{tabular}
\vspace{-6mm}
\end{table}

\vspace{+1mm}
\noindent \textbf{Extension to Category-level SBIR:}
\label{sec:SBIR}
For categorical SBIR we use Sketchy (ext)~\cite{liu2017deep} dataset (90:10 train:test split), which holds $75k$ sketches across $125$ categories having $\approx73k$ images~\cite{liu2017deep} in total. It is evaluated using mean average precision (mAP@all) and precision at top 200 retrieval (P@200) \cite{liu2017deep}. Here we compete against state-of-the-arts (SoTA) like \textit{B-D2S} that follows \cite{dey2019doodle} without its zero-shot setting, and a simple baseline using a Siamese-style VGG-16 network (\textit{B-SBIR-SN}) following \cite{yu2016sketch}. Similar to \Cref{sec:competitors} adapting SoTAs as teachers in our \textit{SketchyNetV2} paradigm (T-SoTA) shows them retaining their accuracy (\Cref{tab:quant_sBIR}) while significantly lowering compute.

\vspace{+1mm}
\noindent \textbf{Teacher-Student Pairs:} To explore other teacher-student pairs apart from ours, we test SketchyNetV2 models of a few ($F_T, F_S$) pairs, where teachers are trained similar to B-VGG16-SN. From \Cref{tab:TS-pairs} we see our pair to outperform others.

\begin{table}[h]
\vspace{-2mm}
\setlength{\tabcolsep}{4pt}
\scriptsize
\centering
\caption{Evaluation on Teacher-Student Pairs}
\vspace{-3mm}
\begin{tabular}{lcccccc}
\toprule
\multirow{2}{*}{\diaghead{AAAAAAAAA}{\tiny{\rotatebox[origin=c]{-21}{Teacher}}}{\tiny{\rotatebox[origin=c]{-21}{Student}}}} 
 & \multicolumn{2}{c}{MobileNetV2} & \multicolumn{2}{c}{EfficientNet} & \multicolumn{2}{c}{ResNet-18} \\
 \cmidrule(lr){2-3} \cmidrule(lr){4-5} \cmidrule(l){6-7}
& Top-1 & FLOPs & Top-1 & FLOPs & Top-1 & FLOPs \\
\cmidrule(r){1-1} \cmidrule(lr){2-3} \cmidrule(lr){4-5} \cmidrule(l){6-7}
 VGG-16 \cite{simonyan2015very}       & \cellcolor{YellowGreen!40}32.77\% & \cellcolor{YellowGreen!40}0.254G & 29.47\% & 0.459G & 26.72\% & 1.451G \\
 Inception-V3 \cite{Yu2015SketchaNetTB} & 27.89\% & 0.264G & 27.16\% & 0.461G & 25.31\% & 1.452G \\
\bottomrule
\end{tabular}
\label{tab:TS-pairs}
\vspace{-2mm}
\end{table}

\vspace{+1mm}
\noindent \textbf{Combining compression methods:} To explore if a combination of compression methods is capable of increasing efficiency, we subject our Knowledge Distillation (KD) trained model (SketchyNetV1) of Triplet-SN \cite{yu2016sketch} (\Cref{tab:quant}) to \textit{(i)} quantisation following \cite{gupta2015deep}, and \textit{(ii)} network-pruning following \cite{molchanov2016pruning}, separately, and then train their respective canvas-selectors. While the former secures $27.34\%$ (\vs $27.89\%$ \textit{ours}) Top-1 score with a lower inference time of $9.1$ ms (\vs $10$ ms \textit{ours}), the latter scores $26.91\%$ (\vs $27.89\%$ ms \textit{ours}) Top-1 with lower FLOPs of $0.249$G (\vs $0.264$G \textit{ours}). Such competitive results further endorse exploring combining compression methods as future works.

\vspace{+1mm}
\noindent \textbf{Future Works:} 
Sketch being an abstraction of an object's photo, when resized to a lower scale for retrieval, infers that its object's semantic is interpretable at that scale. This should ideally provide a lower bound for resizing a photo while retaining interpretability (\textit{e.g.} high retrieval score). Consequently, designing a learnable photo-resizer aided by our trained sketch-canvas selector is a potential future work. Secondly, ours being a meta-framework, using more recent frameworks like DINOv2 \cite{oquab2023dinov2} as large-scale teacher models for further gains is another targeted future work.

\vspace{-2mm}
\section{Conclusion}
\label{sec:conclusion}
\vspace{-2mm}
\noindent 
We for the first time investigate the problem of efficient inference for FG-SBIR, where we show that using existing efficient light-weight networks directly, are unfit for sketches. We thus propose two generic components that can be adapted to photo-networks for sketch-specific data. Firstly, a cross-modal knowledge distillation paradigm distils the knowledge of larger models to a smaller one, directly reducing FLOPs (params) by $97.92$ ($84.9$)\%. Secondly, an abstraction-aware canvas-size selector dynamically selects the best canvas-size for a sketch, reducing FLOPs by extra two-thirds. Extensive experiments show our method to perform at par with state-of-the-arts using much lower compute thus further conveying our method's worth in enhancing them towards deployment.

%
%

\small
\bibliographystyle{ieeenat_fullname}
\bibliography{main}

\begin{thebibliography}{82}
\providecommand{\natexlab}[1]{#1}
\providecommand{\url}[1]{\texttt{#1}}
\expandafter\ifx\csname urlstyle\endcsname\relax
  \providecommand{\doi}[1]{doi: #1}\else
  \providecommand{\doi}{doi: \begingroup \urlstyle{rm}\Url}\fi

\bibitem[Alwani et~al.(2022)Alwani, Wang, and Madhavan]{alwani2022decore}
Manoj Alwani, Yang Wang, and Vashisht Madhavan.
\newblock Decore: Deep compression with reinforcement learning.
\newblock In \emph{CVPR}, 2022.

\bibitem[{Aneeshan Sain and Ayan Kumar Bhunia and Pinaki Nath Chowdhury and Aneeshan Sain and Subhadeep Koley and Tao Xiang and Yi-Zhe Song}(2023)]{sain2023clip}
{Aneeshan Sain and Ayan Kumar Bhunia and Pinaki Nath Chowdhury and Aneeshan Sain and Subhadeep Koley and Tao Xiang and Yi-Zhe Song}.
\newblock {CLIP for All Things Zero-Shot Sketch-Based Image Retrieval, Fine-Grained or Not}.
\newblock In \emph{CVPR}, 2023.

\bibitem[Ba and Caruana(2014)]{ba2014deep}
Jimmy Ba and Rich Caruana.
\newblock Do deep nets really need to be deep?
\newblock \emph{NeurIPS}, 2014.

\bibitem[Bagherinezhad et~al.(2018)Bagherinezhad, Horton, Rastegari, and Farhadi]{bagherinezhad2018label}
Hessam Bagherinezhad, Maxwell Horton, Mohammad Rastegari, and Ali Farhadi.
\newblock Label refinery: Improving imagenet classification through label progression.
\newblock \emph{arXiv preprint arXiv:1805.02641}, 2018.

\bibitem[Bhunia et~al.(2020{\natexlab{a}})Bhunia, Das, Muhammad, Yang, Hospedales, Xiang, Gryaditskaya, and Song]{bhunia2020pixelor}
Ayan~Kumar Bhunia, Ayan Das, Umar~Riaz Muhammad, Yongxin Yang, Timothy~M Hospedales, Tao Xiang, Yulia Gryaditskaya, and Yi-Zhe Song.
\newblock Pixelor: A competitive sketching ai agent. so you think you can sketch?
\newblock \emph{ACM TOG}, 2020{\natexlab{a}}.

\bibitem[Bhunia et~al.(2020{\natexlab{b}})Bhunia, Yang, Hospedales, Xiang, and Song]{bhunia2020sketch}
Ayan~Kumar Bhunia, Yongxin Yang, Timothy~M Hospedales, Tao Xiang, and Yi-Zhe Song.
\newblock Sketch less for more: On-the-fly fine-grained sketch based image retrieval.
\newblock In \emph{CVPR}, 2020{\natexlab{b}}.

\bibitem[Bhunia et~al.(2021{\natexlab{a}})Bhunia, Chowdhury, Sain, Yang, Xiang, and Song]{bhunia2021more}
Ayan~Kumar Bhunia, Pinaki~Nath Chowdhury, Aneeshan Sain, Yongxin Yang, Tao Xiang, and Yi-Zhe Song.
\newblock More photos are all you need: Semi-supervised learning for fine-grained sketch based image retrieval.
\newblock In \emph{CVPR}, 2021{\natexlab{a}}.

\bibitem[Bhunia et~al.(2021{\natexlab{b}})Bhunia, Chowdhury, Yang, Hospedales, Xiang, and Song]{bhunia2021vectorization}
Ayan~Kumar Bhunia, Pinaki~Nath Chowdhury, Yongxin Yang, Timothy~M Hospedales, Tao Xiang, and Yi-Zhe Song.
\newblock Vectorization and rasterization: Self-supervised learning for sketch and handwriting.
\newblock In \emph{CVPR}, 2021{\natexlab{b}}.

\bibitem[Bhunia et~al.(2022)Bhunia, Koley, Khilji, Sain, Chowdhury, Xiang, and Song]{bhunia2022sketching}
Ayan~Kumar Bhunia, Subhadeep Koley, Abdullah Faiz Ur~Rahman Khilji, Aneeshan Sain, Pinaki~Nath Chowdhury, Tao Xiang, and Yi-Zhe Song.
\newblock Sketching without worrying: Noise-tolerant sketch-based image retrieval.
\newblock In \emph{CVPR}, 2022.

\bibitem[Bhunia et~al.(2023)Bhunia, Koley, Kumar, Sain, Chowdhury, Xiang, and Song]{bhunia2023sketch2saliency}
Ayan~Kumar Bhunia, Subhadeep Koley, Amandeep Kumar, Aneeshan Sain, Pinaki~Nath Chowdhury, Tao Xiang, and Yi-Zhe Song.
\newblock {Sketch2Saliency: Learning to Detect Salient Objects from Human Drawings}.
\newblock In \emph{CVPR}, 2023.

\bibitem[Chen et~al.(2020)Chen, Kornblith, Norouzi, and Hinton]{chen2020simple}
Ting Chen, Simon Kornblith, Mohammad Norouzi, and Geoffrey Hinton.
\newblock A simple framework for contrastive learning of visual representations.
\newblock In \emph{ICML}, 2020.

\bibitem[Chen et~al.(2022{\natexlab{a}})Chen, Dai, Chen, Liu, Dong, Yuan, and Liu]{chen2022mobile}
Yinpeng Chen, Xiyang Dai, Dongdong Chen, Mengchen Liu, Xiaoyi Dong, Lu Yuan, and Zicheng Liu.
\newblock Mobile-former: Bridging mobilenet and transformer.
\newblock In \emph{CVPR}, 2022{\natexlab{a}}.

\bibitem[Chen et~al.(2022{\natexlab{b}})Chen, Qiao, Cheng, Pu, Niu, and Li]{chen2022dynamic}
Ying Chen, Liang Qiao, Zhanzhan Cheng, Shiliang Pu, Yi Niu, and Xi Li.
\newblock Dynamic low-resolution distillation for cost-efficient end-to-end text spotting.
\newblock In \emph{ECCV}, 2022{\natexlab{b}}.

\bibitem[Chowdhury et~al.(2022{\natexlab{a}})Chowdhury, Bhunia, Gajjala, Sain, Xiang, and Song]{PartialSBIR}
Pinaki~Nath Chowdhury, Ayan~Kumar Bhunia, Viswanatha~Reddy Gajjala, Aneeshan Sain, Tao Xiang, and Yi-Zhe Song.
\newblock Partially does it: Towards scene-level fg-sbir with partial input.
\newblock In \emph{CVPR}, 2022{\natexlab{a}}.

\bibitem[Chowdhury et~al.(2022{\natexlab{b}})Chowdhury, Bhunia, Gajjala, Sain, Xiang, and Song]{chowdhury2022partially}
Pinaki~Nath Chowdhury, Ayan~Kumar Bhunia, Viswanatha~Reddy Gajjala, Aneeshan Sain, Tao Xiang, and Yi-Zhe Song.
\newblock Partially does it: Towards scene-level fg-sbir with partial input.
\newblock In \emph{The IEEE Conference on Computer Vision and Pattern Recognition (CVPR)}, 2022{\natexlab{b}}.

\bibitem[Chowdhury et~al.(2022{\natexlab{c}})Chowdhury, Sain, Bhunia, Xiang, Gryaditskaya, and Song]{chowdhury2022fs}
Pinaki~Nath Chowdhury, Aneeshan Sain, Ayan~Kumar Bhunia, Tao Xiang, Yulia Gryaditskaya, and Yi-Zhe Song.
\newblock Fs-coco: Towards understanding of freehand sketches of common objects in context.
\newblock In \emph{ECCV}, 2022{\natexlab{c}}.

\bibitem[{Chowdhury, Pinaki Nath and Bhunia, Ayan Kumar and Sain, Aneeshan and Koley, Subhadeep and Xiang, Tao and Song, Yi-Zhe}(2023)]{chowdhury2023democratising}
{Chowdhury, Pinaki Nath and Bhunia, Ayan Kumar and Sain, Aneeshan and Koley, Subhadeep and Xiang, Tao and Song, Yi-Zhe}.
\newblock {Democratising 2D Sketch to 3D Shape Retrieval Through Pivoting}.
\newblock In \emph{ICCV}, 2023.

\bibitem[Collomosse et~al.(2019)Collomosse, Bui, and Jin]{collomosse2019livesketch}
John Collomosse, Tu Bui, and Hailin Jin.
\newblock Livesketch: Query perturbations for guided sketch-based visual search.
\newblock In \emph{CVPR}, 2019.

\bibitem[Courbariaux et~al.(2015)Courbariaux, Bengio, and David]{courbariaux2015binaryconnect}
Matthieu Courbariaux, Yoshua Bengio, and Jean-Pierre David.
\newblock Binaryconnect: Training deep neural networks with binary weights during propagations.
\newblock \emph{NeurIPS}, 28, 2015.

\bibitem[Dey et~al.(2019)Dey, Riba, Dutta, Llados, and Song]{dey2019doodle}
Sounak Dey, Pau Riba, Anjan Dutta, Josep Llados, and Yi-Zhe Song.
\newblock Doodle to search: Practical zero-shot sketch-based image retrieval.
\newblock In \emph{CVPR}, 2019.

\bibitem[Dutta and Akata(2019)]{dutta2019semantically}
Anjan Dutta and Zeynep Akata.
\newblock Semantically tied paired cycle consistency for zero-shot sketch-based image retrieval.
\newblock In \emph{CVPR}, 2019.

\bibitem[Gupta et~al.(2015)Gupta, Agrawal, Gopalakrishnan, and Narayanan]{gupta2015deep}
Suyog Gupta, Ankur Agrawal, Kailash Gopalakrishnan, and Pritish Narayanan.
\newblock Deep learning with limited numerical precision.
\newblock In \emph{ICML}, 2015.

\bibitem[Ha and Eck(2018)]{ha2018neural}
David Ha and Douglas Eck.
\newblock A neural representation of sketch drawings.
\newblock In \emph{ICLR}, 2018.

\bibitem[Han et~al.(2016)Han, Mao, and Dally]{han2015deep}
Song Han, Huizi Mao, and William~J Dally.
\newblock Deep compression: Compressing deep neural networks with pruning, trained quantization and huffman coding.
\newblock In \emph{ICLR}, 2016.

\bibitem[He et~al.(2016)He, Zhang, Ren, and Sun]{HeResNet}
Kaiming He, Xiangyu Zhang, Shaoqing Ren, and Jian Sun.
\newblock Deep residual learning for image recognition.
\newblock In \emph{CVPR}, 2016.

\bibitem[Hinton et~al.(2014)Hinton, Vinyals, and Dean]{hinton2015distilling}
Geoffrey Hinton, Oriol Vinyals, and Jeff Dean.
\newblock Distilling the knowledge in a neural network.
\newblock In \emph{NeurIPS Deep Learning Workshop}, 2014.

\bibitem[Huang et~al.(2017)Huang, Liu, Van Der~Maaten, and Weinberger]{huang2017densely}
Gao Huang, Zhuang Liu, Laurens Van Der~Maaten, and Kilian~Q Weinberger.
\newblock Densely connected convolutional networks.
\newblock In \emph{CVPR}, 2017.

\bibitem[Huang and Wu(2021)]{huang2021robust}
Shouyou Huang and Qiang Wu.
\newblock Robust pairwise learning with huber loss.
\newblock \emph{Journal of Complexity}, 2021.

\bibitem[Huang and Wang(2017)]{huang2017like}
Zehao Huang and Naiyan Wang.
\newblock Like what you like: Knowledge distill via neuron selectivity transfer.
\newblock \emph{arXiv preprint arXiv:1707.01219}, 2017.

\bibitem[Kaelbling et~al.(1996)Kaelbling, Littman, and Moore]{kaelbling1996reinforcement}
Leslie~Pack Kaelbling, Michael~L Littman, and Andrew~W Moore.
\newblock Reinforcement learning: A survey.
\newblock \emph{JAIR}, 1996.

\bibitem[Karen~Simonyan(2015)]{simonyan2015very}
Andrew~Zisserman Karen~Simonyan.
\newblock Very deep convolutional networks for large-scale image recognition.
\newblock In \emph{ICLR}, 2015.

\bibitem[Koley et~al.(2024)Koley, Bhunia, Sain, Chowdhury, Xiang, and Song]{koley2024handle}
Subhadeep Koley, Ayan~Kumar Bhunia, Aneeshan Sain, Pinaki~Nath Chowdhury, Tao Xiang, and Yi-Zhe Song.
\newblock How to handle sketch-abstraction in sketch-based image retrieval?
\newblock In \emph{CVPR}, 2024.

\bibitem[Li et~al.(2018)Li, Wu, Zhang, and Huang]{li2018a2}
Debang Li, Huikai Wu, Junge Zhang, and Kaiqi Huang.
\newblock A2-rl: Aesthetics aware reinforcement learning for image cropping.
\newblock In \emph{CVPR}, 2018.

\bibitem[Li et~al.(2020)Li, Zhang, and Huang]{li2020learning}
Debang Li, Junge Zhang, and Kaiqi Huang.
\newblock Learning to learn cropping models for different aspect ratio requirements.
\newblock In \emph{CVPR}, 2020.

\bibitem[Liang et~al.(2017)Liang, Lee, and Xing]{liang2017deep}
Xiaodan Liang, Lisa Lee, and Eric~P Xing.
\newblock Deep variation-structured reinforcement learning for visual relationship and attribute detection.
\newblock In \emph{CVPR}, 2017.

\bibitem[Lin et~al.(2016)Lin, Talathi, and Annapureddy]{lin2016fixed}
Darryl Lin, Sachin Talathi, and Sreekanth Annapureddy.
\newblock Fixed point quantization of deep convolutional networks.
\newblock In \emph{ICML}, 2016.

\bibitem[Lin et~al.(2020)Lin, Fu, Jiang, and Xue]{lin2020sketch}
Hangyu Lin, Yanwei Fu, Yu-Gang Jiang, and Xiangyang Xue.
\newblock Sketch-bert: Learning sketch bidirectional encoder representation from transformers by self-supervised learning of sketch gestalt.
\newblock In \emph{CVPR}, 2020.

\bibitem[Liu et~al.(2017)Liu, Shen, Shen, Liu, and Shao]{liu2017deep}
Li Liu, Fumin Shen, Yuming Shen, Xianglong Liu, and Ling Shao.
\newblock Deep sketch hashing: Fast free-hand sketch-based image retrieval.
\newblock In \emph{CVPR}, 2017.

\bibitem[Liu et~al.(2023)Liu, Peng, Zheng, Yang, Hu, and Yuan]{liu2023efficientvit}
Xinyu Liu, Houwen Peng, Ningxin Zheng, Yuqing Yang, Han Hu, and Yixuan Yuan.
\newblock Efficientvit: Memory efficient vision transformer with cascaded group attention.
\newblock In \emph{CVPR}, 2023.

\bibitem[Luo et~al.(2017)Luo, Wu, and Lin]{luo2017thinet}
Jian-Hao Luo, Jianxin Wu, and Weiyao Lin.
\newblock Thinet: A filter level pruning method for deep neural network compression.
\newblock In \emph{ICCV}, 2017.

\bibitem[Molchanov et~al.(2016)Molchanov, Tyree, Karras, Aila, and Kautz]{molchanov2016pruning}
Pavlo Molchanov, Stephen Tyree, Tero Karras, Timo Aila, and Jan Kautz.
\newblock Pruning convolutional neural networks for resource efficient inference.
\newblock \emph{arXiv preprint arXiv:1611.06440}, 2016.

\bibitem[Muhammad et~al.(2018)Muhammad, Yang, Song, Xiang, and Hospedales]{muhammad2018learning}
Umar~Riaz Muhammad, Yongxin Yang, Yi-Zhe Song, Tao Xiang, and Timothy~M Hospedales.
\newblock Learning deep sketch abstraction.
\newblock In \emph{CVPR}, 2018.

\bibitem[Muhammad et~al.(2019)Muhammad, Yang, Hospedales, Xiang, and Song]{umar2019goal}
Umar~Riaz Muhammad, Yongxin Yang, Timothy Hospedales, Tao Xiang, and Yi-Zhe Song.
\newblock Goal-driven sequential data abstraction.
\newblock In \emph{ICCV}, 2019.

\bibitem[Oquab et~al.(2023)Oquab, Darcet, Moutakanni, Vo, Szafraniec, Khalidov, Fernandez, Haziza, Massa, El-Nouby, et~al.]{oquab2023dinov2}
Maxime Oquab, Timoth{\'e}e Darcet, Th{\'e}o Moutakanni, Huy Vo, Marc Szafraniec, Vasil Khalidov, Pierre Fernandez, Daniel Haziza, Francisco Massa, Alaaeldin El-Nouby, et~al.
\newblock {DINOv2: Learning Robust Visual Features without Supervision}.
\newblock \emph{arXiv preprint arXiv:2304.07193}, 2023.

\bibitem[Pang et~al.(2017)Pang, Song, Xiang, and Hospedales]{pang2017cross}
Kaiyue Pang, Yi-Zhe Song, Tony Xiang, and Timothy~M Hospedales.
\newblock Cross-domain generative learning for fine-grained sketch-based image retrieval.
\newblock In \emph{BMVC}, 2017.

\bibitem[Pang et~al.(2019)Pang, Li, Yang, Zhang, Hospedales, Xiang, and Song]{pang2019generalising}
Kaiyue Pang, Ke Li, Yongxin Yang, Honggang Zhang, Timothy~M Hospedales, Tao Xiang, and Yi-Zhe Song.
\newblock Generalising fine-grained sketch-based image retrieval.
\newblock In \emph{CVPR}, 2019.

\bibitem[Pang et~al.(2020)Pang, Yang, Hospedales, Xiang, and Song]{pang2020solving}
Kaiyue Pang, Yongxin Yang, Timothy~M Hospedales, Tao Xiang, and Yi-Zhe Song.
\newblock Solving mixed-modal jigsaw puzzle for fine-grained sketch-based image retrieval.
\newblock In \emph{CVPR}, 2020.

\bibitem[Passalis and Tefas(2018)]{passalis2018learning}
Nikolaos Passalis and Anastasios Tefas.
\newblock Learning deep representations with probabilistic knowledge transfer.
\newblock In \emph{ECCV}, 2018.

\bibitem[{Pinaki Nath Chowdhury and Ayan Kumar Bhunia and Aneeshan Sain and Subhadeep Koley and Tao Xiang and Yi-Zhe Song}(2023{\natexlab{a}})]{chowdhury2023scenetrilogy}
{Pinaki Nath Chowdhury and Ayan Kumar Bhunia and Aneeshan Sain and Subhadeep Koley and Tao Xiang and Yi-Zhe Song}.
\newblock {SceneTrilogy: On Human Scene-Sketch and its Complementarity with Photo and Text}.
\newblock In \emph{CVPR}, 2023{\natexlab{a}}.

\bibitem[{Pinaki Nath Chowdhury and Ayan Kumar Bhunia and Aneeshan Sain and Subhadeep Koley and Tao Xiang and Yi-Zhe Song}(2023{\natexlab{b}})]{chowdhury2023what}
{Pinaki Nath Chowdhury and Ayan Kumar Bhunia and Aneeshan Sain and Subhadeep Koley and Tao Xiang and Yi-Zhe Song}.
\newblock {What Can Human Sketches Do for Object Detection?}
\newblock In \emph{CVPR}, 2023{\natexlab{b}}.

\bibitem[Romero et~al.(2015)Romero, Ballas, Kahou, Chassang, Gatta, and Bengio]{romero2015fitnets}
Adriana Romero, Nicolas Ballas, Samira~Ebrahimi Kahou, Antonie Chassang, Carlo Gatta, and Yoshua Bengio.
\newblock Fitnets: Hints for thin deep nets.
\newblock In \emph{ICLR}, 2015.

\bibitem[Russakovsky et~al.(2015)Russakovsky, Deng, Su, Krause, Satheesh, Ma, Huang, Karpathy, Khosla, Bernstein, et~al.]{russakovsky2015imagenet}
Olga Russakovsky, Jia Deng, Hao Su, Jonathan Krause, Sanjeev Satheesh, Sean Ma, Zhiheng Huang, Andrej Karpathy, Aditya Khosla, Michael Bernstein, et~al.
\newblock Imagenet large scale visual recognition challenge.
\newblock \emph{IJCV}, 2015.

\bibitem[Sain et~al.(2020)Sain, Bhunia, Yang, Xiang, and Song]{sain2020cross}
Aneeshan Sain, Ayan~Kumar Bhunia, Yongxin Yang, Tao Xiang, and Yi-Zhe Song.
\newblock Cross-modal hierarchical modelling forfine-grained sketch based image retrieval.
\newblock In \emph{BMVC}, 2020.

\bibitem[Sain et~al.(2021)Sain, Bhunia, Yang, Xiang, and Song]{sain2021stylemeup}
Aneeshan Sain, Ayan~Kumar Bhunia, Yongxin Yang, Tao Xiang, and Yi-Zhe Song.
\newblock Stylemeup: Towards style-agnostic sketch-based image retrieval.
\newblock In \emph{CVPR}, 2021.

\bibitem[Sain et~al.(2022)Sain, Bhunia, Potlapalli, Chowdhury, Xiang, and Song]{sketch3T}
Aneeshan Sain, Ayan~Kumar Bhunia, Vaishnav Potlapalli, Pinaki~Nath Chowdhury, Tao Xiang, and Yi-Zhe Song.
\newblock Sketch3t: Test-time training for zero-shot sbir.
\newblock In \emph{CVPR}, 2022.

\bibitem[Sain et~al.(2023)Sain, Bhunia, Koley, Chowdhury, Chattopadhyay, Xiang, and Song]{sain2023exploiting}
Aneeshan Sain, Ayan~Kumar Bhunia, Subhadeep Koley, Pinaki~Nath Chowdhury, Soumitri Chattopadhyay, Tao Xiang, and Yi-Zhe Song.
\newblock {Exploiting Unlabelled Photos for Stronger Fine-Grained SBIR}.
\newblock In \emph{CVPR}, 2023.

\bibitem[Sandler et~al.(2018)Sandler, Howard, Zhu, Zhmoginov, and Chen]{sandler2018mobilenetv2}
Mark Sandler, Andrew Howard, Menglong Zhu, Andrey Zhmoginov, and Liang-Chieh Chen.
\newblock Mobilenetv2: Inverted residuals and linear bottlenecks.
\newblock In \emph{CVPR}, 2018.

\bibitem[Sangkloy et~al.(2016)Sangkloy, Burnell, Ham, and Hays]{sangkloy2016sketchy}
Patsorn Sangkloy, Nathan Burnell, Cusuh Ham, and James Hays.
\newblock The sketchy database: learning to retrieve badly drawn bunnies.
\newblock \emph{ACM TOG}, 2016.

\bibitem[Schulman et~al.(2017)Schulman, Wolski, Dhariwal, Radford, and Klimov]{schulman2017proximal}
John Schulman, Filip Wolski, Prafulla Dhariwal, Alec Radford, and Oleg Klimov.
\newblock Proximal policy optimization algorithms.
\newblock \emph{arXiv preprint arXiv:1707.06347}, 2017.

\bibitem[Simonyan and Zisserman(2015)]{SimonyanZ14a}
Karen Simonyan and Andrew Zisserman.
\newblock Very deep convolutional networks for large-scale image recognition.
\newblock In \emph{ICLR}, 2015.

\bibitem[Song et~al.(2017{\natexlab{a}})Song, Song, Xiang, and Hospedales]{song2017fine}
Jifei Song, Yi-Zhe Song, Tony Xiang, and Timothy~M Hospedales.
\newblock Fine-grained image retrieval: the text/sketch input dilemma.
\newblock In \emph{BMVC}, 2017{\natexlab{a}}.

\bibitem[Song et~al.(2017{\natexlab{b}})Song, Yu, Song, Xiang, and Hospedales]{song2017deep}
Jifei Song, Qian Yu, Yi-Zhe Song, Tao Xiang, and Timothy~M Hospedales.
\newblock Deep spatial-semantic attention for fine-grained sketch-based image retrieval.
\newblock In \emph{ICCV}, 2017{\natexlab{b}}.

\bibitem[{Subhadeep Koley and Ayan Kumar Bhunia and Aneeshan Sain and Pinaki Nath Chowdhury and Tao Xiang and Yi-Zhe Song}(2023)]{koley2023picture}
{Subhadeep Koley and Ayan Kumar Bhunia and Aneeshan Sain and Pinaki Nath Chowdhury and Tao Xiang and Yi-Zhe Song}.
\newblock {Picture that Sketch: Photorealistic Image Generation from Abstract Sketches}.
\newblock In \emph{CVPR}, 2023.

\bibitem[Sutton et~al.(2000)Sutton, McAllester, Singh, and Mansour]{sutton2000policy}
Richard~S Sutton, David~A McAllester, Satinder~P Singh, and Yishay Mansour.
\newblock Policy gradient methods for reinforcement learning with function approximation.
\newblock In \emph{NeurIPS}, 2000.

\bibitem[Talebi and Milanfar(2021)]{talebi2021learning}
Hossein Talebi and Peyman Milanfar.
\newblock Learning to resize images for computer vision tasks.
\newblock In \emph{ICCV}, 2021.

\bibitem[Tan and Le(2019)]{tan2019efficientnet}
Mingxing Tan and Quoc Le.
\newblock Efficientnet: Rethinking model scaling for convolutional neural networks.
\newblock In \emph{ICML}, 2019.

\bibitem[Uzkent and Ermon(2020)]{uzkent2020learning}
Burak Uzkent and Stefano Ermon.
\newblock Learning when and where to zoom with deep reinforcement learning.
\newblock In \emph{CVPR}, 2020.

\bibitem[Vatathanavaro et~al.(2018)Vatathanavaro, Tungjitnob, and Pasupa]{vatathanavaro2018white}
Supawit Vatathanavaro, Suchat Tungjitnob, and Kitsuchart Pasupa.
\newblock White blood cell classification: a comparison between vgg-16 and resnet-50 models.
\newblock In \emph{JSCI}, 2018.

\bibitem[Visvalingam and Whyatt(1990)]{visvalingam1990douglas}
Mahes Visvalingam and J~Duncan Whyatt.
\newblock The douglas-peucker algorithm for line simplification: re-evaluation through visualization.
\newblock In \emph{Computer Graphics Forum}, 1990.

\bibitem[Wang et~al.(2021)Wang, Sapkota, Liu, and Yu]{wang2021deep}
Dingrong Wang, Hitesh Sapkota, Xumin Liu, and Qi Yu.
\newblock Deep reinforced attention regression for partial sketch based image retrieval.
\newblock In \emph{ICDM}, 2021.

\bibitem[Wang et~al.(2019)Wang, Huang, Celikyilmaz, Gao, Shen, Wang, Wang, and Zhang]{wang2019reinforced}
Xin Wang, Qiuyuan Huang, Asli Celikyilmaz, Jianfeng Gao, Dinghan Shen, Yuan-Fang Wang, William~Yang Wang, and Lei Zhang.
\newblock Reinforced cross-modal matching and self-supervised imitation learning for vision-language navigation.
\newblock In \emph{CVPR}, 2019.

\bibitem[Wang et~al.(2016)Wang, Xu, You, Tao, and Xu]{wang2016cnnpack}
Yunhe Wang, Chang Xu, Shan You, Dacheng Tao, and Chao Xu.
\newblock Cnnpack: Packing convolutional neural networks in the frequency domain.
\newblock \emph{NeurIPS}, 29, 2016.

\bibitem[Wang et~al.(2020)Wang, Lv, Huang, Song, Yang, and Huang]{wang2020glance}
Yulin Wang, Kangchen Lv, Rui Huang, Shiji Song, Le Yang, and Gao Huang.
\newblock Glance and focus: a dynamic approach to reducing spatial redundancy in image classification.
\newblock In \emph{NeurIPS}, 2020.

\bibitem[Weinberger and Saul(2009)]{weinberger2009distance}
Kilian~Q Weinberger and Lawrence~K Saul.
\newblock Distance metric learning for large margin nearest neighbor classification.
\newblock \emph{JMLR}, 2009.

\bibitem[Xu et~al.(2021)Xu, Joshi, and Bresson]{xu2021multigraph}
Peng Xu, Chaitanya~K Joshi, and Xavier Bresson.
\newblock Multigraph transformer for free-hand sketch recognition.
\newblock \emph{IEEE T-NNLS}, 2021.

\bibitem[Xu et~al.(2022{\natexlab{a}})Xu, Hospedales, Yin, Song, Xiang, and Wang]{xu2022deep}
Peng Xu, Timothy~M Hospedales, Qiyue Yin, Yi-Zhe Song, Tao Xiang, and Liang Wang.
\newblock Deep learning for free-hand sketch: A survey.
\newblock \emph{IEEE TPAMI}, 2022{\natexlab{a}}.

\bibitem[Xu et~al.(2022{\natexlab{b}})Xu, Mu, Lee, Mukherjee, Chaterji, Bagchi, and Li]{xu2022smartadapt}
Ran Xu, Fangzhou Mu, Jayoung Lee, Preeti Mukherjee, Somali Chaterji, Saurabh Bagchi, and Yin Li.
\newblock {SmartAdapt}: Multi-branch object detection framework for videos on mobiles.
\newblock In \emph{CVPR}, 2022{\natexlab{b}}.

\bibitem[Yu et~al.(2015)Yu, Yang, Song, Xiang, and Hospedales]{Yu2015SketchaNetTB}
Qian Yu, Yongxin Yang, Yi-Zhe Song, Tao Xiang, and Timothy~M. Hospedales.
\newblock Sketch-a-net that beats humans.
\newblock In \emph{BMVC}, 2015.

\bibitem[Yu et~al.(2016)Yu, Liu, Song, Xiang, Hospedales, and Loy]{yu2016sketch}
Qian Yu, Feng Liu, Yi-Zhe Song, Tao Xiang, Timothy~M Hospedales, and Chen-Change Loy.
\newblock Sketch me that shoe.
\newblock In \emph{CVPR}, 2016.

\bibitem[Zagoruyko and Komodakis(2017)]{zagoruyko2017KDattention}
Sergey Zagoruyko and Nikos Komodakis.
\newblock {Paying more attention to attention: Improving the performance of convolutional neural networks via attention transfer}.
\newblock In \emph{ICLR}, 2017.

\bibitem[Zhou et~al.(2016)Zhou, Wu, Ni, Zhou, Wen, and Zou]{zhou2016dorefa}
Shuchang Zhou, Yuxin Wu, Zekun Ni, Xinyu Zhou, He Wen, and Yuheng Zou.
\newblock Dorefa-net: Training low bitwidth convolutional neural networks with low bitwidth gradients.
\newblock \emph{arXiv preprint arXiv:1606.06160}, 2016.

\bibitem[Zhu et~al.(2021)Zhu, Han, Wu, Zhang, Nie, Lan, and Wang]{zhu2021dynamic}
Mingjian Zhu, Kai Han, Enhua Wu, Qiulin Zhang, Ying Nie, Zhenzhong Lan, and Yunhe Wang.
\newblock Dynamic resolution network.
\newblock In \emph{NeurIPS}, 2021.

\end{thebibliography}

\newpage

\twocolumn[
  \begin{@twocolumnfalse}
    \begin{center}\textbf{\Large{
    Supplementary Material for\\
    \vspace{2mm}
    Sketch Down the FLOPs: Towards Efficient Networks for Human Sketch
    }}\end{center}
  \end{@twocolumnfalse}
]

\setcounter{table}{0}
\renewcommand{\thetable}{\Alph{table}}

\section*{Further analyses:}

{
\vspace{-4mm}
\setlength{\tabcolsep}{4.5pt}
\begin{table}[ht]
\caption{Performance \& compute for Transformer-based networks}
\vspace{-2mm}
\scriptsize
\centering
\begin{tabular}{l|ccccc}
\toprule
{Model (on 224$\times$224)}          & {Swin-B} & {PVT-L} & {ViT-B/16} & {DeiT-B/16} & {CvT-21} \\
{FLOPs (approx.)}     & 15.4G & 9.8G & 17.8G & 17.6G & 9.2G \\
{Params (approx.)}    & 88M & 61.4M & 86M & 87M & 32M\\
{Top-1 (ShoeV2)} & 40.71\% & 44.18\% & 16.28\% & 35.62\% & 41.58\%\\
\bottomrule
\end{tabular}
\label{tab:model_flops_params_TF}
\end{table}
}

{
\vspace{-4mm}
\setlength{\tabcolsep}{4pt}
\renewcommand{\arraystretch}{0.95}
\begin{table}[h]
    \centering
    \scriptsize
    \caption{Results on transformer-based architectures.}
    \vspace{-2mm}
    \begin{tabular}{lccc}
    \toprule
    Model & Self (\%) & SketchyNetV1 (\%) & SketchyNetV2 (\%) \\
    \cmidrule(lr){1-4}
    SketchPVT \cite{sain2023exploiting} &  44.18 & 43.01 & 42.47 \\
    Swin-B \cite{sain2023exploiting} &  40.71 & 38.94 & 38.02 \\
    EfficientViT-M5 \cite{liu2023efficientvit} & 38.76 & 37.29 & 36.58 \\
    \cmidrule(lr){1-4}
    SketchAbstract \cite{koley2024handle} & 45.31 & 44.06 & 43.59 \\  
    Partial \cite{wang2021deep} (full-sketch) & 32.87 & 31.33 & 30.79 \\
    \cmidrule(lr){1-4}
    ViT \cite{sain2023exploiting} / MobileNetV2 & 16.28 & 12.53 & 11.64 \\
    ViT \cite{sain2023exploiting} / MobileFormer & 16.28 & 14.86 & 13.65 \\
    \bottomrule
\end{tabular}
\label{tab:transformer_results}
\end{table}
}

\noindent\textbf{Generalization to transformers:}
\Cref{tab:model_flops_params_TF} shows the Top-1 score on ShoeV2 of popular transformers (trained like \cite{sain2023exploiting}), and also experiment (Top-1 on ShoeV2) using suggested methods (from \cite{sain2023exploiting}) as teachers and MobileNetV2 as student. To recap, the aim is to \textit{retain the accuracy of the teacher model} in its smaller variants of SketchyNetV1 and SketchyNetV2. Two more experiments use the same teacher of pretrained Vit (trained on Triplet Loss), to train \textit{(a)} MobileFormer \cite{chen2022mobile} (transformer) and \textit{(b)} MobileNetV2 (CNN).
Tabulated results (\Cref{tab:transformer_results}) show SketchyNetV2 variants to hold nearly similar accuracy as their teachers, akin to our findings in Table~\red{2}, thus validating our method. 

\noindent \textbf{Other ablations and experiments:}
\textit{(i)} While \cite{wang2021deep} uses deep reinforced attention regression to tackle unnecessary strokes for retrieval from \textit{partial} sketches and \cite{sain2023exploiting} explores transformers for accuracy our goal is to deliver the \textit{same} accuracy of a large model at a much \textit{reduced compute} thus addressing the \textit{sparsity} in a sketch. 
\textit{(ii)} On varying $\lambda$ as (0.1, 0.15, $\cdots$ 0.9), accuracy falls when $\lambda$ $>$ 0.7 or $\lambda$ $<$ 0.45, residing optimally at $\lambda$ = 0.5, thus suggesting an equal impact of triplet loss and KD component.

\noindent\textbf{Distinguishing our work from~\cite{sain2023exploiting} for clarity:}
While our work might share certain components with \cite{sain2023exploiting}, it fundamentally differs in motivation, methodology, and novelties:
\textit{(i)} ours is the \textbf{first} investigation into efficient sketch networks adapted from photo-based ones.
\textit{(ii)} Unlike \cite{sain2023exploiting}, which applies KD on fixed-resolution sketches, our novel abstraction-aware canvas-size selector dynamically optimizes sketch resolutions, achieving a 97.92\% reduction in FLOPs while retaining accuracy.
\textit{(iii)} Our KD framework uniquely uses sketch features from the teacher, always extracted at \textit{full} resolution, to guide the student operating on \textit{varying} resolutions, invoking \textbf{scale invariance}, a property ignored in \cite{sain2023exploiting}, and crucial for our scenario.
\textit{(iv)} We specifically address the sparsity and abstraction inherent to sketches, while \cite{sain2023exploiting} focuses primarily on dense visual representations.

\noindent \textbf{Miscellaneous:}
\textit{(i)} Compared to MSE or MAE losses, which are prone to noisy outliers, \textbf{Huber loss} provides robustness by balancing convergence and stability \cite{huang2021robust}. It enhances generalization, reducing sensitivity to small deviations and large outliers, particularly benefiting our cross-modal FGSBIR task.
\textit{(ii)} Interestingly, more sketches were downsized for ShoeV2 (22.31\% at 128x128, 38.12\% to 64x64, 17.64\% to 32x32 ) compared to FSCOCO (16.75\% at 128x128, 1.2\% to 64x64) dataset, likely because the latter holds much more detailed sketches.

\end{document}